\documentclass{IEEEoj}
\usepackage{cite}
\usepackage{amsmath,amssymb,amsfonts}
\usepackage{algorithmic}
\usepackage{graphicx,color}
\usepackage{textcomp}
\usepackage{url}            % simple URL typesetting
\usepackage{booktabs}       % professional-quality tables
\usepackage{amsfonts}       % blackboard math symbols
\usepackage{nicefrac}       % compact symbols for 1/2, etc.
\usepackage{microtype}      % microtypography
\usepackage{amsmath}
\usepackage{graphicx}       % graphics
\usepackage{subcaption}
\usepackage{multirow}

\def\BibTeX{{\rm B\kern-.05em{\sc i\kern-.025em b}\kern-.08em
    T\kern-.1667em\lower.7ex\hbox{E}\kern-.125emX}}
\AtBeginDocument{\definecolor{ojcolor}{cmyk}{0.93,0.59,0.15,0.02}}
\def\OJlogo{\vspace{-14pt}\includegraphics[height=28pt]{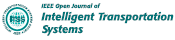}}
\begin{document}
\receiveddate{XX Month, XXXX}
\reviseddate{XX Month, XXXX}
\accepteddate{XX Month, XXXX}
\publisheddate{XX Month, XXXX}
\currentdate{XX Month, XXXX}
\doiinfo{OJITS.2022.1234567}

\title{DST-TransitNet: A Dynamic Spatio-Temporal Deep Learning Model for Scalable and Efficient Network-Wide Prediction of Station-Level Transit Ridership}

\author{Jiahao Wang\authorrefmark{1}, Amer Shalaby\authorrefmark{1}}
\affil{Department of Civil and Mineral Engineering (Transportation Engineering), University of Toronto, Toronto, Ontario M5S 1A4 Canada}
\corresp{CORRESPONDING AUTHOR: Jiahao Wang (e-mail: jhope.wang@mail.utoronto.ca).}
% \authornote{This work was supported by the Natural Sciences and Engineering Research Council (NSERC) of Canada.}
\markboth{Preparation of Papers for IEEE Transactions on Intelligent Transportation Systems}{Author \textit{Wang and Shalaby}}

\begin{abstract}
Accurate prediction of public transit ridership is vital for efficient planning and management of transit in rapidly growing urban areas in Canada. Unexpected increases in passengers can cause overcrowded vehicles, longer boarding times, and service disruptions. Traditional time series models like ARIMA and SARIMA face limitations, particularly in short-term predictions and integration of spatial and temporal features. These models struggle with the dynamic nature of ridership patterns and often ignore spatial correlations between nearby stops.Deep Learning (DL) models present a promising alternative, demonstrating superior performance in short-term prediction tasks by effectively capturing both spatial and temporal features. However, challenges such as dynamic spatial feature extraction, balancing accuracy with computational efficiency, and ensuring scalability remain.

This paper introduces DST-TransitNet, a hybrid DL model for system-wide station-level ridership prediction. This proposed model uses graph neural networks (GNN) and recurrent neural networks (RNN) to dynamically integrate the changing temporal and spatial correlations within the stations. The model also employs a precise time series decomposition framework to enhance accuracy and interpretability. Tested on Bogotá’s BRT system data, with three distinct social scenarios, DST-TransitNet outperformed state-of-the-art models in precision, efficiency and robustness. Meanwhile, it maintains stability over long prediction intervals, demonstrating practical applicability.
\end{abstract}



\maketitle

\section{INTRODUCTION} \IEEEPARstart{A}{ccurate} is crucial for efficient planning and management of transit in the context of urban development in Canada. Cities in Canada are experiencing significant population growth, leading to higher demand for public transit services. Unanticipated spikes in transit ridership can result in overcrowded vehicles, longer boarding and alighting times, and operational issues such as bus bunching. These problems can accumulate and propagate through the transit network, causing widespread service disruptions. Accurate prediction of transit ridership enables transit agencies to optimize resource allocation, adjust vehicle deployment, and schedule frequencies to match expected rider levels.

Traditional time series models, such as ARIMA and SARIMA, are widely used for similar prediction tasks. However, they face challenges and limitations in predicting transit ridership for stops in the transit network. Firstly, traditional methods struggle to provide accurate results for the prediction of short-term transit ridership. Although parametric methods such as ARIMA are suitable for long-term prediction intervals, such as one-day predictions, they falter in short-term prediction tasks due to the rapidly changing dynamics of transit-rider patterns, resulting in lower prediction accuracy. Secondly, traditional methods have difficulty combining spatial and temporal features for predictions. The prediction of transit ridership is often performed for individual stations, neglecting the spatial correlation features. For short-term predictions, considering stops or stations as basic units, similar riding patterns can be shared among nearby or connected stops. Although such correlations can provide additional information for prediction, traditional models struggle to incorporate spatial and temporal features simultaneously because of their structural limitations.

A promising solution to these challenges is the use of Deep Learning (DL) models. Research indicates that DL models perform better on short-term time series prediction tasks in various applications, such as traffic flow prediction and stock prediction~\cite{boukerche2020machine,redhu2023short}. DL models can capture subtle pattern patterns in the system when trained with sufficient data. In addition, they demonstrate efficiency in station-level predictions by capturing spatial and temporal features through ensembles of multiple models. For example, Convolutional Neural Networks (CNNs) can capture spatial features, while Recurrent Neural Networks (RNNs) can capture temporal features.

However, several challenges must be addressed before applying DL structures to the prediction of short-term transit ridership. Firstly, temporal-spatial feature extraction should not only consider static spatial features but also incorporate dynamic aspects. Although the overall transit network may remain unchanged, the correlation levels between the linked stations / stops can vary throughout the day. Secondly, the prediction model must balance accuracy with computational efficiency. Although complex model structures capture data set details, training and deployment of deep learning models require substantial computational resources, which may be prohibitive for some public transit organizations. Third, the model should demonstrate good scalability, adapting to dataset changes and maintaining accuracy in new situations, such as significant social events or pandemics. From a practical perspective, the model should have not only higher resolution but also longer prediction period, (i.e. it should perform well on both short and long prediction intervals).

In this paper, we present our work on the prediction of short-term transit ridership using deep learning (DL) models to address the challenges mentioned above. Our proposed model, DST-TransitNet (Dynamic Spatio-Temporal Transit Network), integrates temporal and spatial features cohesively and dynamically by employing a highly integrated Graph Neural Network (GNN) and Recurrent Neural Network (RNN). In addition, the model incorporates a more precise and explicable time series decomposition framework to improve the accuracy and interpretability of the prediction. The model was tested on a dataset of real-world public transit ridership collected from the Bogotá Rapid Transit (BRT) system of Bogotá, Colombia~\footnote{https://datosabiertos-transmilenio.hub.arcgis.com}, and compared with state-of-the-art machine learning (ML) models in terms of accuracy, stability, and efficiency. Our experiments demonstrate that DST-TransitNet exhibits accurate, efficient, and robust performance in network-wide station-level ridership prediction tasks across various prediction scenarios. Furthermore, the model shows excellent stability over longer prediction intervals, significantly increasing its practical application value.

The rest of the paper is organized as follows: In the next section, we provide a general review of the literature on transit demand prediction tasks and state-of-the-art time series prediction models. Next, we present the proposed model for short-term passenger demand prediction and explain the framework we applied for making use of short-term prediction model for longer-term prediction tasks. Subsequently, we introduce the dataset used, delve into the experiment results, provide a detailed analysis of the prediction results on different real-world scenarios and different prediction lengths, and we further discuss the effects of different dataset patterns on prediction outcomes. In the final section, we offer conclusions and discuss potential future work.

\section{Background}\label{sec:background}
Public transit ridership analysis has evolved significantly, ranging from statistical modeling to machine learning and deep learning methods for time-series prediction. This literature review discusses various ridership analysis approaches across multiple dimensions, such as problem definitions, model features, and methodologies.

Statistical analysis is one of the most common approaches for ridership estimation, including methods like direct ridership modeling and travel demand models~\cite{deepa2022direct}. Direct rider modeling focuses on assessing how different factors within a buffer range, such as socioeconomic conditions, land use, variables of the built environment, and attributes related to transport, affect transport demand~\cite{cervero2010direct, gutierrez2011transit}. For example, \cite{jun2015land} uses characteristics of land use, such as population and employment density, to estimate the demand for subway stations in the Seoul metropolitan area, further analyzing the impact of these factors in different reference areas. Similarly, \cite{vergel2018ridership} explores the relationship between the built environment and the ridership in the bus rapid transit (BRT) systems in seven Latin American cities. In Montreal, infrastructure conditions, along with characteristics of the built environment, are examined to estimate bus ridership at the stop level~\cite{chakour2016examining}. Traditional direct ridership models focus mainly on long-term ridership estimation, ranging from monthly~\cite{gutierrez2011transit} to daily forecasts~\cite{liu2016increase}.

Traditional statistical models are heavily based on selected factors. Physical attributes, such as density, diversity, and design, alongside social variables, are primarily long-term characteristics that are useful for strategic planning or policy making, particularly in forecasting ridership for proposed transit routes. However, these features do not capture real-time impacts on ridership. To address the need for short-term estimations, real-time factors such as weather conditions or events are incorporated into ridership models~\cite{stover2012impact, liu2016measuring}. For example, \cite{santanam2024public} examines the impact of planned and unplanned events on transit ridership using a linear regression and random forest-based model.

Another factor influencing the performance of statistical analysis models is their structure. Regression models, such as ordinary least squares (OLS), are widely used due to their interpretability, showing relationships between dependent and independent variables~\cite{cardozo2012application, thompson2012really, zhao2013relationship}. However, as time resolution decreases and more real-time features are added, traditional linear regression models face issues such as multicollinearity and struggle with more complex attributes~\cite{ding2016predicting}. To address these limitations, more advanced models are employed. For example, \cite{tao2018travel} uses a SARIMAX model to estimate the hourly bus capacity by incorporating weather conditions. \cite{chayan2024predicting} employs a hybrid agent-based model that integrates factors of land use and travel demand for accurate estimation of short-term transit demand. Machine learning (ML) models have also gained popularity for handling complex input features. \cite{yang2023time} uses a random forest model to improve traditional ridership prediction by capturing non-linear and temporal features of factors of the built environment. A study on the effectiveness of different ML models, including k closest neighbors, random forest, and gradient boost, was carried out for the estimation of the ridership in the 30-minute interval in~\cite{wang2022effectiveness}.

Despite the widespread use of statistical models, they have limitations due to their dependency on factor-based estimation. These models are costly to update, often requiring census data collection, which restricts their adaptability to rapid developments or sudden changes, such as the COVID-19 outbreak.

To address these challenges, time series prediction models are utilized for predicting transit ridership. Unlike statistical models, time series models focus on identifying patterns within historical data. These models learn from historical passenger records, which are frequently updated and collected directly from transit systems, allowing easy deployment and rapid updates~\cite{caicedo2023public}.

Various models are used for time series-based ridership prediction. \cite{milenkovic2018sarima} deploys SARIMA for monthly system-wide ridership prediction using ten years of past ridership data. \cite{ding2017using} proposes an ARIMA model combined with a generalized autoregressive conditional heteroskedasticity (GARCH) approach for short-term (15-minute interval) ridership forecasting. In addition to traditional parametric time-series models, nonparametric models, such as ML and deep learning (DL) models, are also commonly used for time-series-based ridership prediction. \cite{liu2019deeppf} employs a Long Short-Term Memory (LSTM) model for predicting 10-minute interval ridership, while \cite{cheng2022intelligent} uses LSTM to predict next-day passenger counts based on the data from the past seven days. \cite{sha2020rnn} conducts comprehensive experiments to evaluate the performance of various RNN models across different prediction intervals, from short-term (15 minutes) to relatively long-term (6 hours).

Time-series models are easily adaptable for predicting ridership at different geographic scales, where input data can be aggregated to the appropriate level of granularity. Large-scale predictions provide a general understanding of overall system performance, helping to make long-term decisions and system design~\cite{karnberger2020network}. In contrast, smaller-scale predictions help to understand passenger origin-destination flows or station-level crowding, facilitating precise operational decisions~\cite{ji2015transit, hu2020crowding}.

In addition to temporal features, spatial correlation can be extracted from input sequences to capture more connection dimensions between stations. Many studies have focused on combining temporal and spatial features to improve prediction accuracy~\cite{liyanage2022ai, chen2022bidirectional}. \cite{zhao2017lstm} represents spatial relationships between stations using a contributing coefficient matrix (ODC) calculated between historical input sequences and the prediction target from other stations. This approach uses a conventional LSTM model with ODC as its input for final ridership prediction. Another method involves treating the transit network as an image, where each pixel represents normalized ridership values, and using a Convolutional Neural Network (CNN) to extract spatial correlations~\cite{du2019deep}. Further advancements leverage graph neural networks (GNNs) to model spatio-temporal correlations, representing transit networks as graphs using adjacency matrices~\cite{li2020graph, wang2022effective, wu2023learning}. However, considering only the physical connections between stations in graph-based models is insufficient; the temporal similarities between stations must also be accounted for, which remains underexplored in current research. Moreover, the relationships between stops are not static but fluctuate throughout the day, making it essential to dynamically model the connectivity and weights between stops to capture these variations accurately.

While deep learning-based time series prediction models are powerful, they require substantial computing resources, making them costly to deploy and maintain, especially for large-scale networks. Moreover, practical applications demand high accuracy for both short-term and long-term predictions, providing valuable information for decision-making with frequent updates~\cite{di2024machine}.

To address these gaps, this work proposes a system-wide, station-level ridership prediction model with short prediction intervals. The model aims to capture dynamic temporal-spatial correlations within a transit system, where station connections are determined by both physical proximity and dynamically changing temporal similarities. The proposed model also considers efficiency for large-scale network prediction and multiple interval forecasts, providing a more robust, accurate, and applicable prediction approach.

\section{Methodology}\label{sec:method}
In this section, we provide a detailed introduction to the Deep Learning (DL) structure, Dynamic Spatio-Temporal TransitNet (DST-TransitNet), designed to address this task by integrating advanced neural network components to capture both temporal and spatial dependencies in transit demand data. This model leverages Graph Convolutional Neural Networks (GCNs), Graph Attention Networks (GATs), Gated Recurrent Units (GRUs), and Feed Forward Neural Networks (FFNNs) to provide a robust and comprehensive framework for short-term transit ridership prediction.

\subsection{Basic Building Blocks}
In this section, we introduce the fundamental machine learning units used to build DST-TransitNet, including GRU, GCN, GNN, and GAT. These components are crucial for understanding the structure of DST-TransitNet and how each part functions cohesively within the model.

\subsubsection{Gated Recurrent Unit (GRU)}
\begin{figure}[!htbp]
    \centering
    \includegraphics[width=\linewidth]{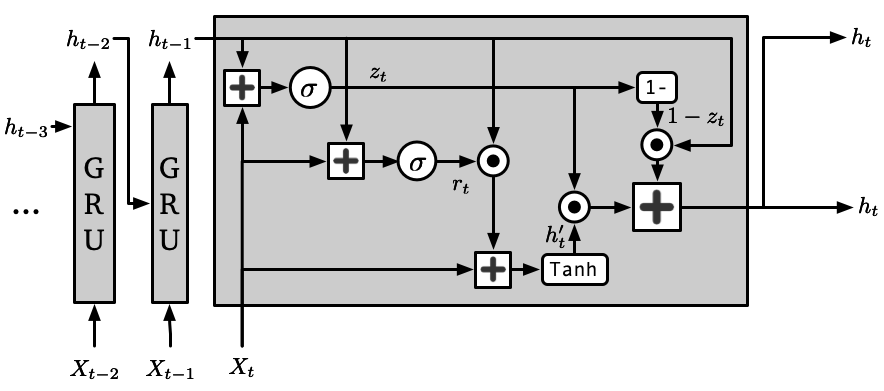} % Change to actual image path
    \caption{Unit structure of a GRU, showing how previous hidden states and current inputs are combined to produce the output at each timestep.}
    \label{fig:gru_structure}
\end{figure}

The GRU, proposed by \cite{chung2014empirical}, is an improvement over traditional RNNs, designed to mitigate the problem of vanishing / exploding gradients. The architecture of the GRU is shown in Figure \ref{fig:gru_structure}, which illustrates how data flows within the unit and how the information is propagated through its structure. As depicted, the hidden state from previous timesteps \(h_{t-1}\) and the current input \(X_t\) are used to calculate the current hidden state \(h_t\). This process captures temporal dependencies across time steps, while the gated structure controls how much information from the previous timestep and current input is retained.

The temporal dependencies at timestep \(t\) are captured by the GRU as follows:

\[
\mathbf{h}_t = (1 - \mathbf{z}_t) \odot \mathbf{h}_{t-1} + \mathbf{z}_t \odot \mathbf{\tilde{h}}_t,
\]
where \( \mathbf{\tilde{h}}_t \) is the candidate activation:
\[
\mathbf{\tilde{h}}_t = \tanh(\mathbf{W}_h \mathbf{x}_t + \mathbf{r}_t \odot (\mathbf{U}_h \mathbf{h}_{t-1}) + \mathbf{b}_h),
\]
and \( \mathbf{z}_t \) and \( \mathbf{r}_t \) are the update and reset gates, respectively:
\[
\begin{aligned}
\mathbf{z}_t &= \sigma(\mathbf{W}_z \mathbf{x}_t + \mathbf{U}_z \mathbf{h}_{t-1} + \mathbf{b}_z), \\
\mathbf{r}_t &= \sigma(\mathbf{W}_r \mathbf{x}_t + \mathbf{U}_r \mathbf{h}_{t-1} + \mathbf{b}_r).
\end{aligned}
\]

\subsubsection{Graph Convolutional Networks}
\begin{figure}
    \centering
    \includegraphics[width=0.5\linewidth]{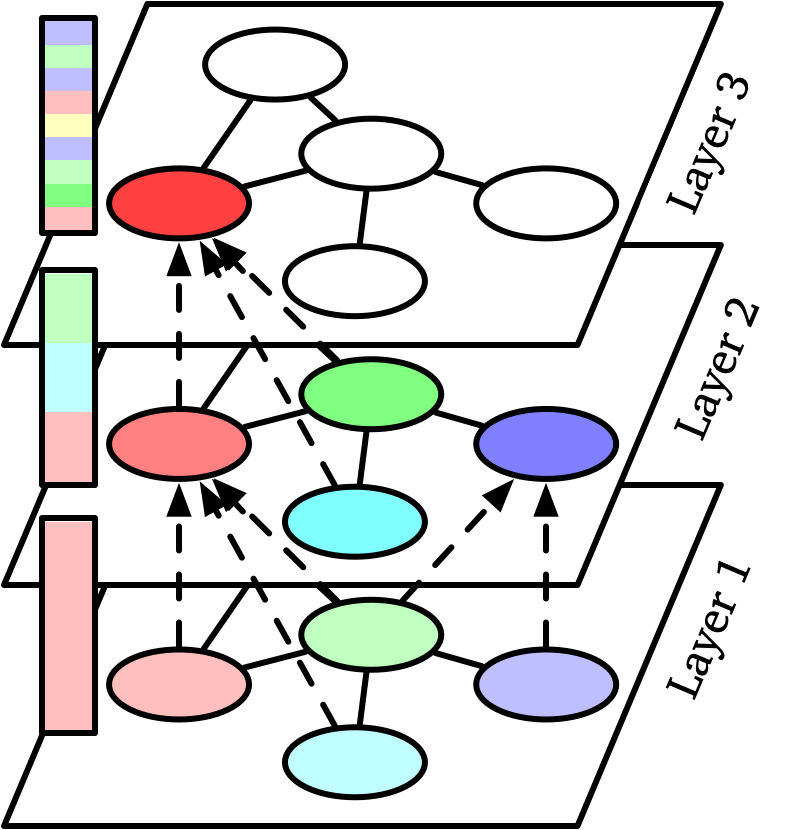}
    \caption{Message passing through a graph-structured dataset via GCN layers}
    \label{fig:gcn_cal}
\end{figure}

Proposed in~\cite{kipf2016semi}, GCN address the limitations of traditional Convolutional Neural Networks (CNNs) in capturing the message or feature passing between nodes in a graph-based data structure. Unlike CNNs, which only allow messages to be passed in two dimensions (as in grid-like structures such as images), GCNs enable message passing between all adjacent nodes in a graph. As shown in Fig.~\ref{fig:gcn_cal}, the feature aggregation for a node is based on messages from all its neighboring nodes. The aggregation is computed using the following equation

\[
H^{(l+1)} = \sigma \left( \tilde{D}^{-\frac{1}{2}} \tilde{A} \tilde{D}^{-\frac{1}{2}} H^{(l)} W^{(l)} \right),
\]
where 
\begin{itemize}
    \item \( H^{(l)} \) represents the feature matrix at the \(l\)-th layer of the network, where each row corresponds to a node and each column corresponds to a feature.
    \item \( W^{(l)} \) is the trainable weight matrix at layer \(l\).
    \item \( \tilde{A} = A + I \) is the adjacency matrix of the graph \( A \) with added identity matrix \( I \), representing self-loops \( I \).
    \item \( \tilde{D} \) is the diagonal degree matrix of \( \tilde{A} \), and \(  \tilde{D}^{-\frac{1}{2}} \tilde{A} \tilde{D}^{-\frac{1}{2}} \) is used for normalize the adjacency matrix.
    \item \( \sigma \) is the activation function.
\end{itemize}

This formulation allows GCNs to capture and aggregate information from neighboring nodes effectively, taking into account the graph structure. The more GCN layer included within the computation, the further infoarmiton can be reached by a given node, with the normalization using \(\tilde{D}\) preventing the feature values from growing disproportionately as the layers stack deeper.

On the other hands, in~\cite{morris2019weisfeiler}, the authors provided a improved version of GCN, named k-dimensional Graph Neural Network (k-GNN), where the message aggregation within a layer is calculated as follow:

\[
H^{(l+1)}(v) = \sigma \left( H^{(l)}(v) \cdot W_1^{(l+1)} + \sum_{w \in A} H^{(l)}(w) \cdot W_2^{(l+1)} \right),
\]
which, compared to traditional GCN, allows both connected and unconnected nodes to communicate with each other.

\subsubsection{Graph Attention Network}
GATs, introduced by \cite{velickovic2018graph}, extend GCNs by incorporating attention mechanisms, allowing dynamic weighting of the neighbors' contributions based on the input. As shown in Fig.~\ref{fig:gat_unit}, GATs calculate dynamic edge weights \(W_E\) based on the attention mechanism, which is formulated as:

\begin{figure}[!htbp]
    \centering
    \includegraphics[width=\linewidth]{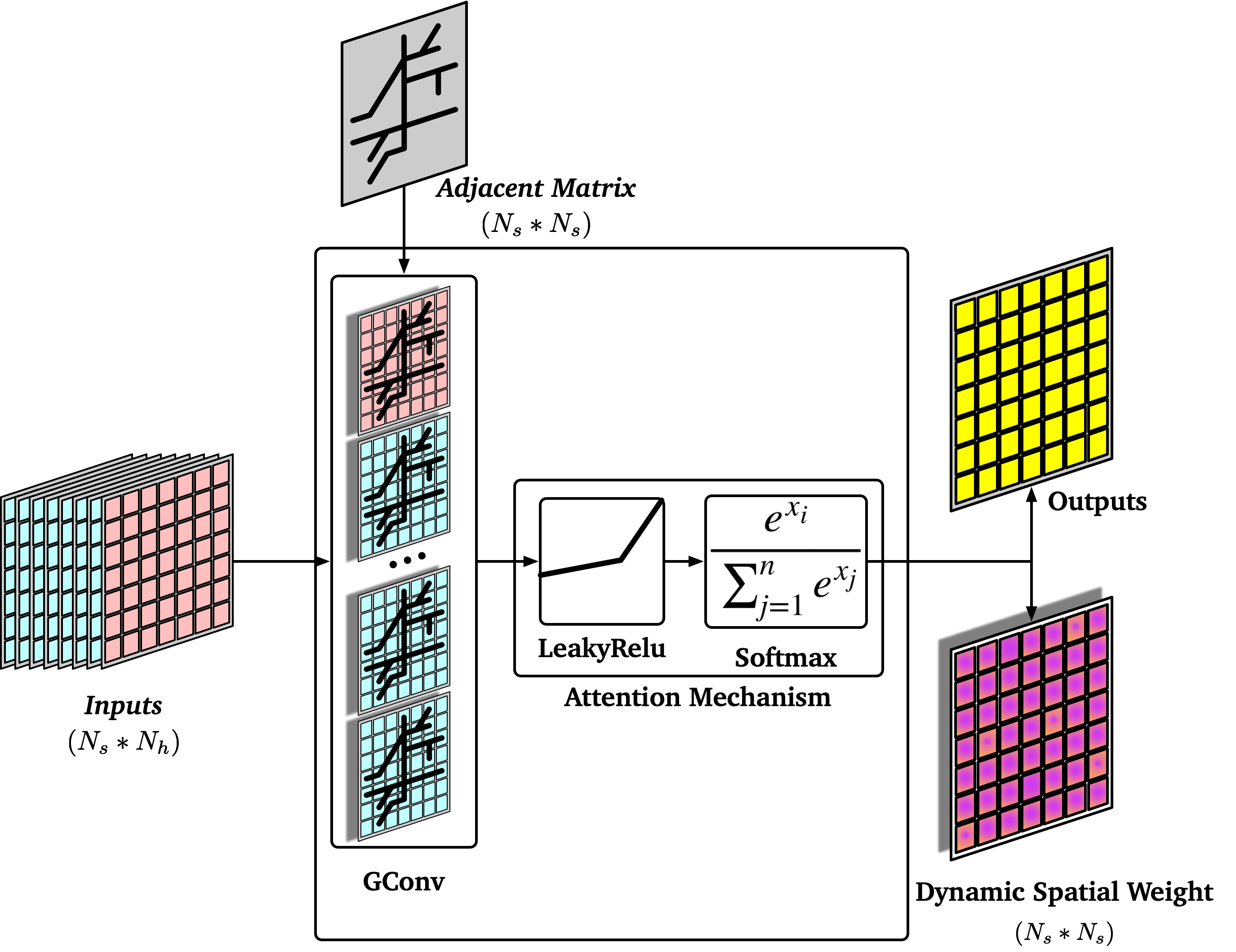}
    \caption{Graph Attention Network (GAT) unit, demonstrating how attention mechanisms are used to weight the contributions of neighboring nodes dynamically.}
    \label{fig:gat_unit}
\end{figure}

\[
e_{ij}^{(l)} = \text{LeakyReLU} \left( \mathbf{a}^{(l)T} \left[ z_i^{(l)} || z_j^{(l)} \right] \right),
\]
where:
\begin{itemize}
    \item \(z_i^{(l)} = \mathbf{W}^{(l)} \mathbf{h}_i^{(l)}\) is the transformed feature vector for node \(i\).
    \item \( || \) denotes concatenation.
    \item \( \text{LeakyReLU} \) is the Leaky Rectified Linear Unit activation function.
\end{itemize}

The edge weights are computed by applying the softmax function to the attention coefficients:
\[
W_{E_{ij}}^{(l)} = \frac{\exp(e_{ij}^{(l)})}{\sum_{k \in \mathcal{N}(i)} \exp(e_{ik}^{(l)})}.
\]
Finally, the new hidden state \( \mathbf{h}_i^{(l+1)} \) is calculated as a weighted sum of the neighbors' transformed features:
\[
\mathbf{h}_i^{(l+1)} = \sigma \left( \sum_{j \in \mathcal{N}(i)} W_{E_{ij}}^{(l)} z_j^{(l)} \right).
\]
The LeakyReLU function is defined as:
\[
\text{LeakyReLU}(x) = 
\begin{cases} 
x, & \text{if } x > 0, \\
\alpha x, & \text{if } x \leq 0,
\end{cases}
\]
where \( \alpha \) is a small constant (typically \( \alpha = 0.01 \)).

GAT enables the model to dynamically focus on the most relevant spatial features for each stop, which is crucial for capturing varying transit ridership across different network areas throughout different time periods.

\subsection{DST-TransitNets}
\begin{figure}[!htbp]
    \centering
    \includegraphics[width=1\linewidth]{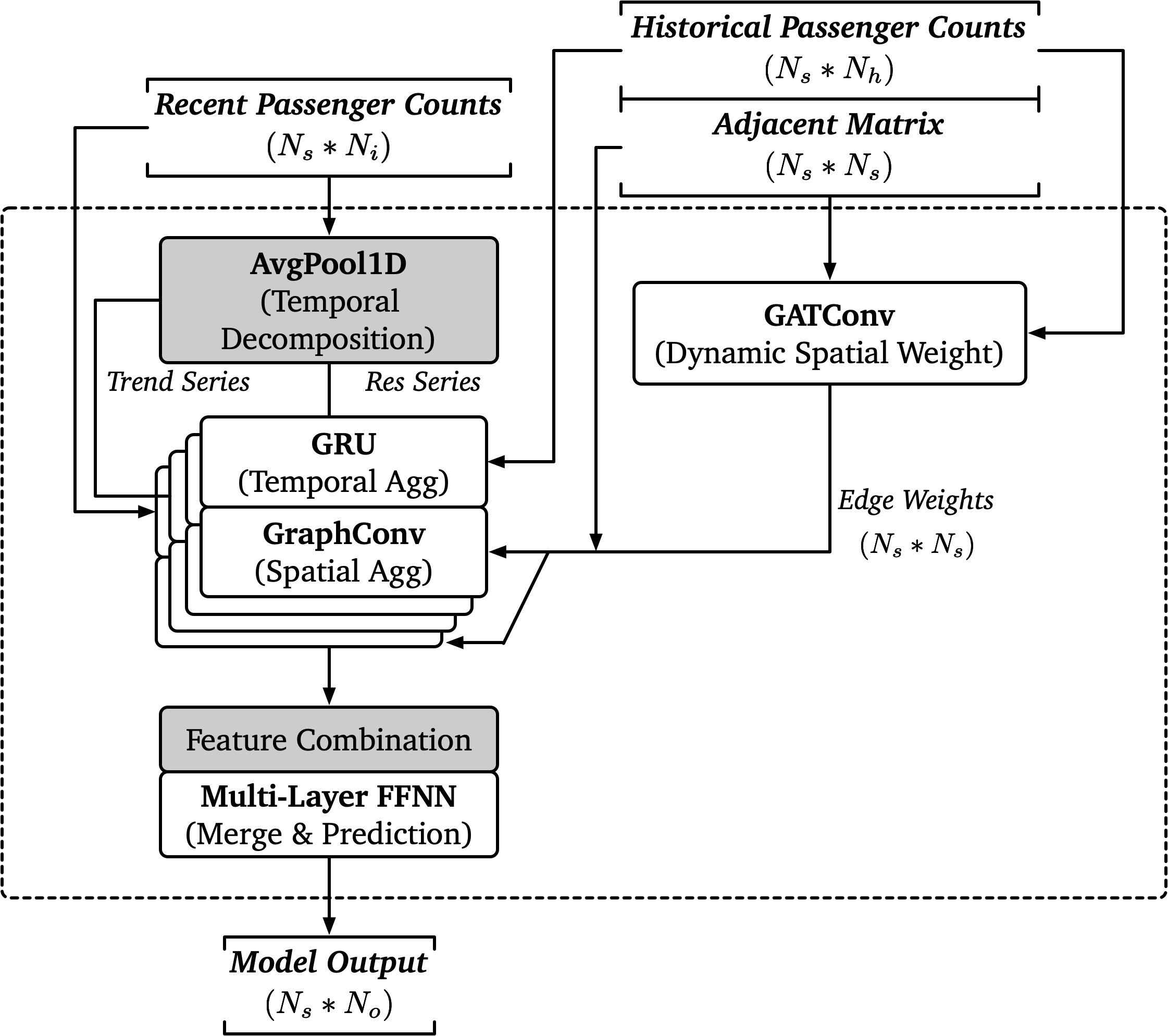}
    \caption{Dynamic Spatio-Temporal TransitNet (DST-TransitNet) Model Structure}
    \label{fig:strc_stlinearv7}
\end{figure}

The structure of the proposed DL model, i.e., DST-TransitNets, is shown in Fig.~\ref{fig:strc_stlinearv7}. The model comprises four main parts: the temporal decomposition layer, dynamic spatial weight calculation layer, spatio-temporal aggregation layer, and prediction layer. The proposed model is capable of utilizing information from all stations in the transit network, such as current passenger counts, historical passenger counts, and spatial relations between stations, to make predictions for all stops/stations simultaneously.

\subsubsection{Temporal Decomposition}
This layer serves as the temporal decomposition component, a structure proven to be effective for time series prediction tasks in many works~\cite{zeng2023transformers, wu2021autoformer}. The main idea of temporal decomposition is to summarize the input data \( X_o \) over a fixed temporal window using average pooling, and then divide the original time series input into Trend series \( X_t \) and Residual series \( X_r \), calculated as follows:

\[\begin{aligned}
        X_t &= AvgPool1D(X_o), \\
    X_r &= X_o - X_t.
\end{aligned}
\]
This layer helps extract key temporal features from the original data, making the data more interpretable and manageable for subsequent layers.

\subsubsection{Dynamic Spatial Weight Calculation}
This layer employs the GAT to calculate dynamic spatial weighting \( \mathbf{W}_E \). Although the transit network structure remains stable, the correlation within stops/stations may change dynamically. GAT assigns different weights to the edges connected to each target node using historical passenger counts.

In our work, GAT dynamically assigns weights \( W_{E} \) to each node \( i \) and its neighbors \( j \) using historical passenger counts information \( X_h \), collected at the same prediction time on the same day of the past week.

We structured the historical passenger counts for GAT as \( \mathbf{X_h} = \{\mathbf{X_h}_1, \mathbf{X_h}_2, \ldots, \mathbf{X_h}_{N_s}\} \), where \( \mathbf{X_h}_i \in \mathbb{R}^{N_h} \) are the input features for \( N_s \) stops, each with \( N_h \) historical records. The attention coefficients to each node \( i \) and its neighbors \( j \) can be calculated as follows:

\[
(e, W_{E}) = \text{GATConv}(\mathbf{X_h}, \mathbf{a}| , \mathbf{w})
\]
where \( \mathbf{a} \in \mathbb{R}^{2N_s} \) is the adjacency matrix, \( \mathbf{w} \in \mathbb{R}^{N_s \times N_h} \) is the trainable layer weight matrix.

\subsubsection{Spatio-Temporal Aggregation}
\begin{figure}[!htbp]
    \centering
    \includegraphics[width=1\linewidth]{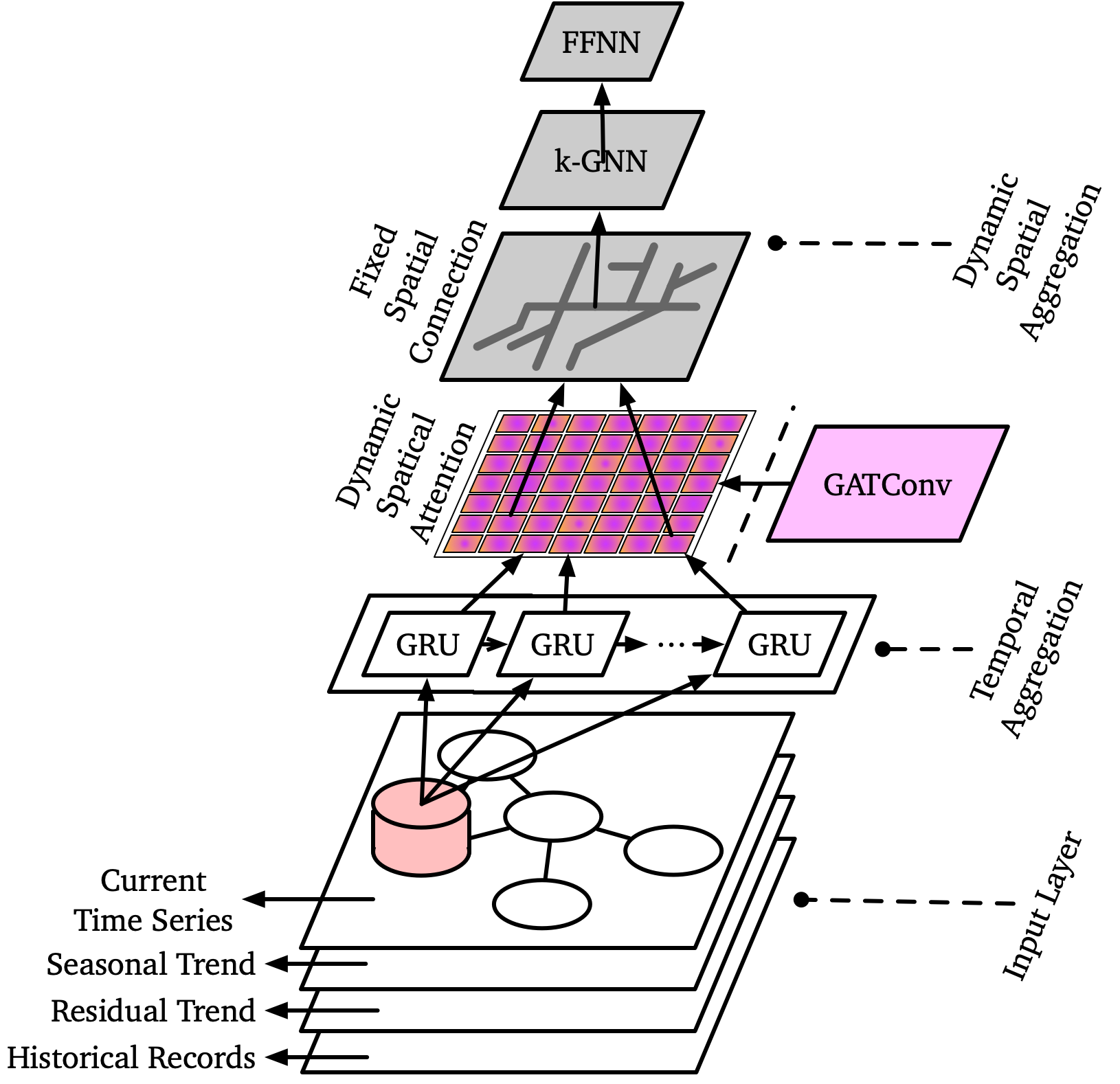}
    \caption{The architecture of the Spatio-Temporal Aggregation layer in DST-TransitNet, combining Dynamic Spatial Aggregation (via k-GNN and GATConv) and Temporal Aggregation (via stacked GRUs).}
    \label{fig:sta_structure}
\end{figure}

The spatio-temporal aggregation layer combines stacked GRUs for temporal feature aggregation and k-GNNs for spatial feature aggregation. This approach processes current passenger counts \( X_o \), historical passenger counts \( X_h \), trend series \( X_t \), and residual series \( X_r \), allowing DST-TransitNet to extract both temporal and spatial dependencies. The detailed architecture of the spatio-temporal aggregation phase in DST-TransitNet is illustrated in Fig.~\ref{fig:sta_structure}.

% The temporal dependencies for timestamp \( t \) are captured by GRU as hidden state \( h_t \):

% \begin{equation}\label{eq:gru1}
%    \mathbf{h}_t = (1 - \mathbf{z}_t) \odot \mathbf{h}_{t-1} + \mathbf{z}_t \odot \mathbf{\tilde{h}}_t,
% \end{equation}
% where $\mathbf{\tilde{h}}_t$ is the candidate activation vector:
% \begin{equation}\label{eq:gru2}
%    \mathbf{\tilde{h}}_t = \tanh(\mathbf{W}_h \mathbf{x}_t + \mathbf{r}_t \odot (\mathbf{U}_h \mathbf{h}_{t-1}) + \mathbf{b}_h).
% \end{equation}
% In Equ.~\ref{eq:gru1} and Equ.~\ref{eq:gru2}, \( \mathbf{z}_t \) is the update gate value, and \( \mathbf{r}_t \) is the reset gate value:
% \begin{equation}
%     \begin{aligned}
%         \mathbf{z}_t &= \sigma(\mathbf{W}_z \mathbf{x}_t + \mathbf{U}_z \mathbf{h}_{t-1} + \mathbf{b}_z),\\
%        \mathbf{r}_t &= \sigma(\mathbf{W}_r \mathbf{x}_t + \mathbf{U}_r \mathbf{h}_{t-1} + \mathbf{b}_r).
%     \end{aligned}
% \end{equation}
Each of the four series candidates—original input, seasonal component, trend component, and correlation—is processed by a GRU to generate hidden states, followed by a graph convolution operation:

\[
h_i = \text{GRU}(\text{series\_agg}_i(\text{temp\_inputs}_i)) 
\]

\[
\begin{aligned}
        h_i &= \text{k-GNN}(h_i, a, {W_{E_{j,i}}} ) \\
    &= \mathbf{W}_1 \mathbf{x}_i + \mathbf{W}_2
    \sum_{j \in \mathcal{N}(i)} {W_{E_{j,i}}} \cdot \mathbf{x}_j,
\end{aligned}
\]
where \( \mathbf{W}_E \) is the edge weights attention information from the dynamic spatial weight calculation layer.

After concatenating the features extracted from the temporal and spatial aggregation layers, the multi-layer FFNN merges the combined features and maps the high-level features to the output predictions.

The proposed hybrid model structure combines temporal and spatial data by leveraging the strengths of multiple neural network architectures, such as GRU for temporal analysis and k-GNN and GAT for spatial analysis. The integration of attention mechanisms through GAT enables the model to dynamically focus on the most relevant spatial features. Finally, the multi-layer FFNN ensures the combined features are effectively utilized for prediction.

\begin{figure}[!htpb]
    \centering
    \includegraphics[width=1\linewidth]{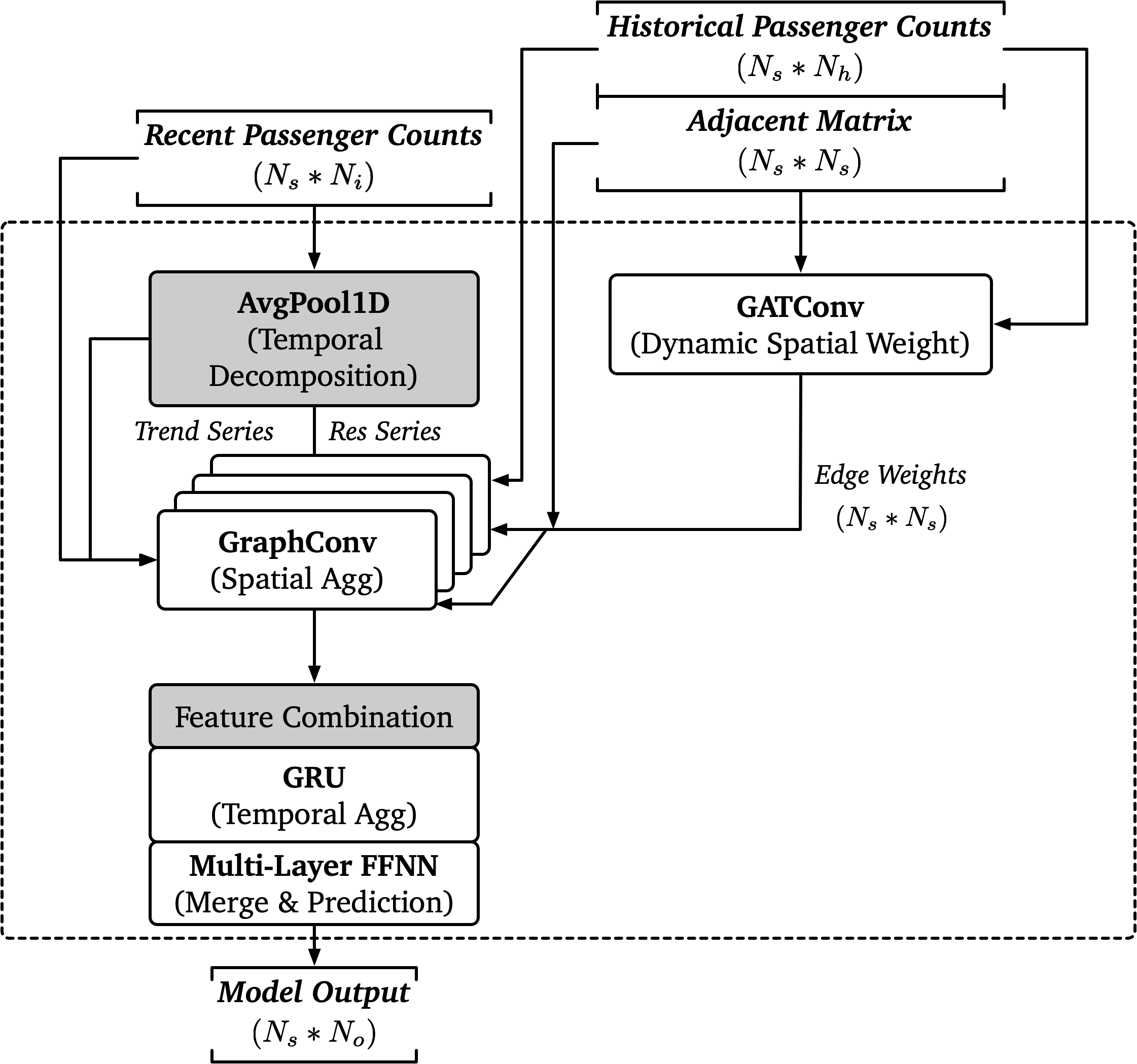}
    \caption{Dynamic Spatio-Temporal TransitNet Version 2 (DST-TransitNetV2) Model Structure}
    \label{fig:strc_stlinearv6}
\end{figure}

To further improve model efficiency, we also provide a second hybrid model, DST-TransitNetV2. In this version, the GRU layer in the original spatio-temporal aggregation layer is moved to the prediction layer, where it processes the spatially-aggregated features for high-level temporal feature extraction, as shown in Fig.~\ref{fig:strc_stlinearv6}.

\subsubsection{Long-term Prediction Using Short-term Prediction Model} \label{sc:ltst}
In this section, we describe the iterative framework used to extend a short-term prediction model for long-term transit ridership forecasting. This approach leverages the pre-trained short-term prediction model to dynamically generate extended predictions while maintaining computational efficiency and adaptability.

\begin{figure}[htbp]
    \centering
    \includegraphics[width=1\linewidth]{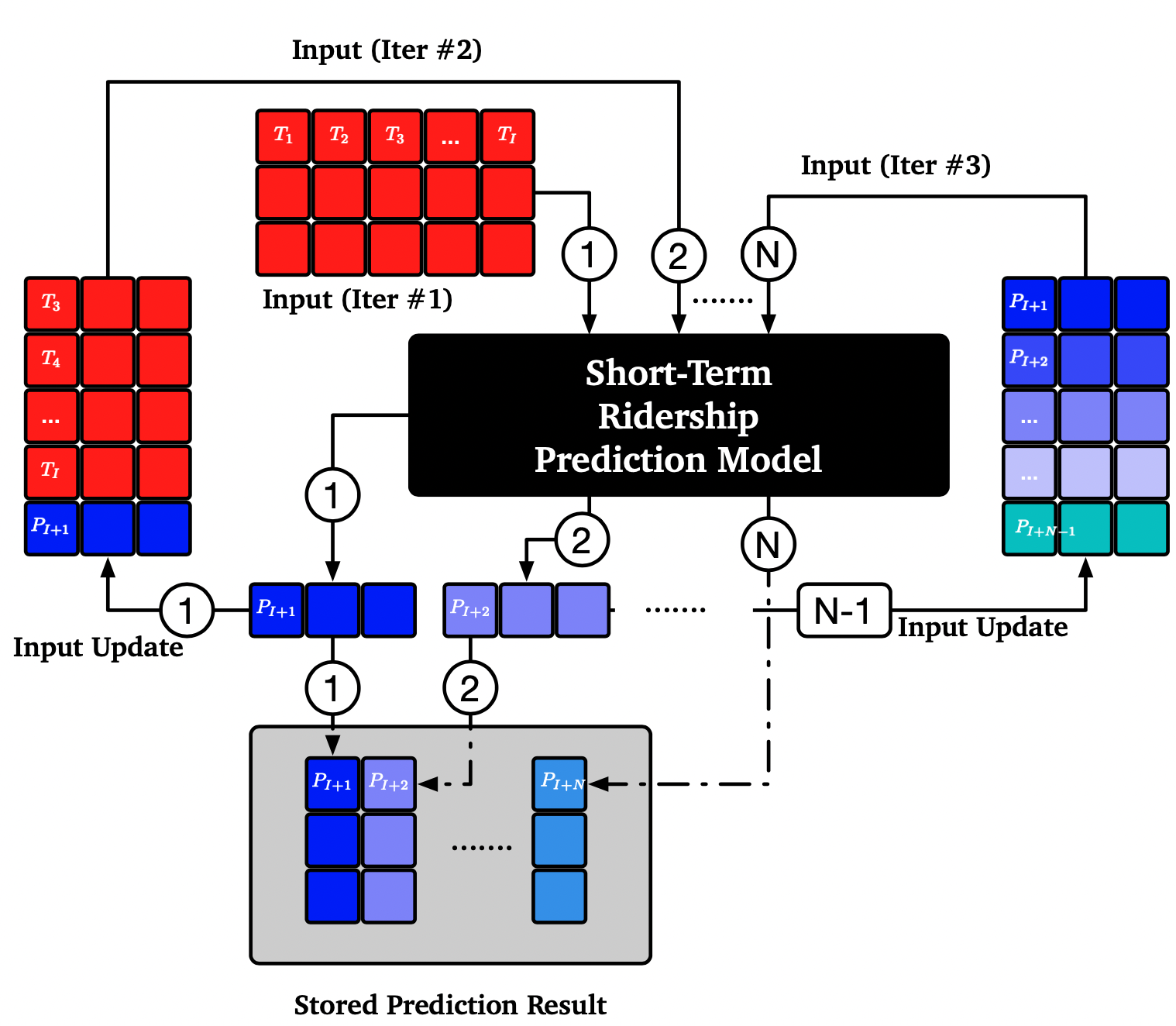}
    \caption{Illustration of the iterative process for extending short-term predictions to long-term forecasts.}
    \label{fig:long_with_short}
\end{figure}

The Fig.~\ref{fig:long_with_short} illustrates the prediction process where $T_i$, represents the original input data at time step $i$, and $P_j$ represents the prediction results for future timestamps $j$. The process involves the following steps:
\begin{enumerate}
    \item Initial Inputs:
        \begin{itemize}
            \item The model begins with an initial set of input data points, denoted as \( T_1, T_2, \ldots, T_I \).
        \end{itemize}
    \item First Iteration (Iter \#1):
        \begin{itemize}
            \item The short-term ridership prediction model processes these initial inputs to generate the first future prediction, \( P_{I+1} \).
        \end{itemize}
    \item Subsequent Iterations (Iter \#2 and beyond):
        \begin{itemize}
            \item In each subsequent iteration, the model updates its input data by incorporating the latest prediction and removing the oldest input. For example, in the second iteration, the inputs become \( T_2, T_3, \ldots, P_{I+1} \) to predict \( P_{I+2} \). This process continues iteratively for \( N \) iterations.
        \end{itemize}
    \item Prediction Storage:
        \begin{itemize}
            \item The predictions from each iteration are stored sequentially, creating a comprehensive long-term prediction sequence from \( P_{I+1} \) to \( P_{I+N} \).
        \end{itemize}
\end{enumerate}

By continuously incorporating new predictions into the input set, the framework enables the model to generate long-term predictions efficiently. This iterative approach reuses the model's structure and parameters in each iteration, thereby avoiding extensive retraining and minimizing the computational resources required for long-term predictions. Such efficiency and scalability are particularly beneficial for public transit agencies with limited computational capacities. However, this framework imposes strict requirements on the model's robustness and accuracy, as long-term predictions increasingly depend on the model's own predictions rather than actual observed data. The iterative nature of the framework can lead to error propagation, where inaccuracies in early predictions compound over subsequent iterations, potentially degrading the accuracy of long-term forecasts.

\subsection{Pipeline for Real-World Deployment}

\begin{figure}[!htbp]
    \centering
    \includegraphics[width=\linewidth]{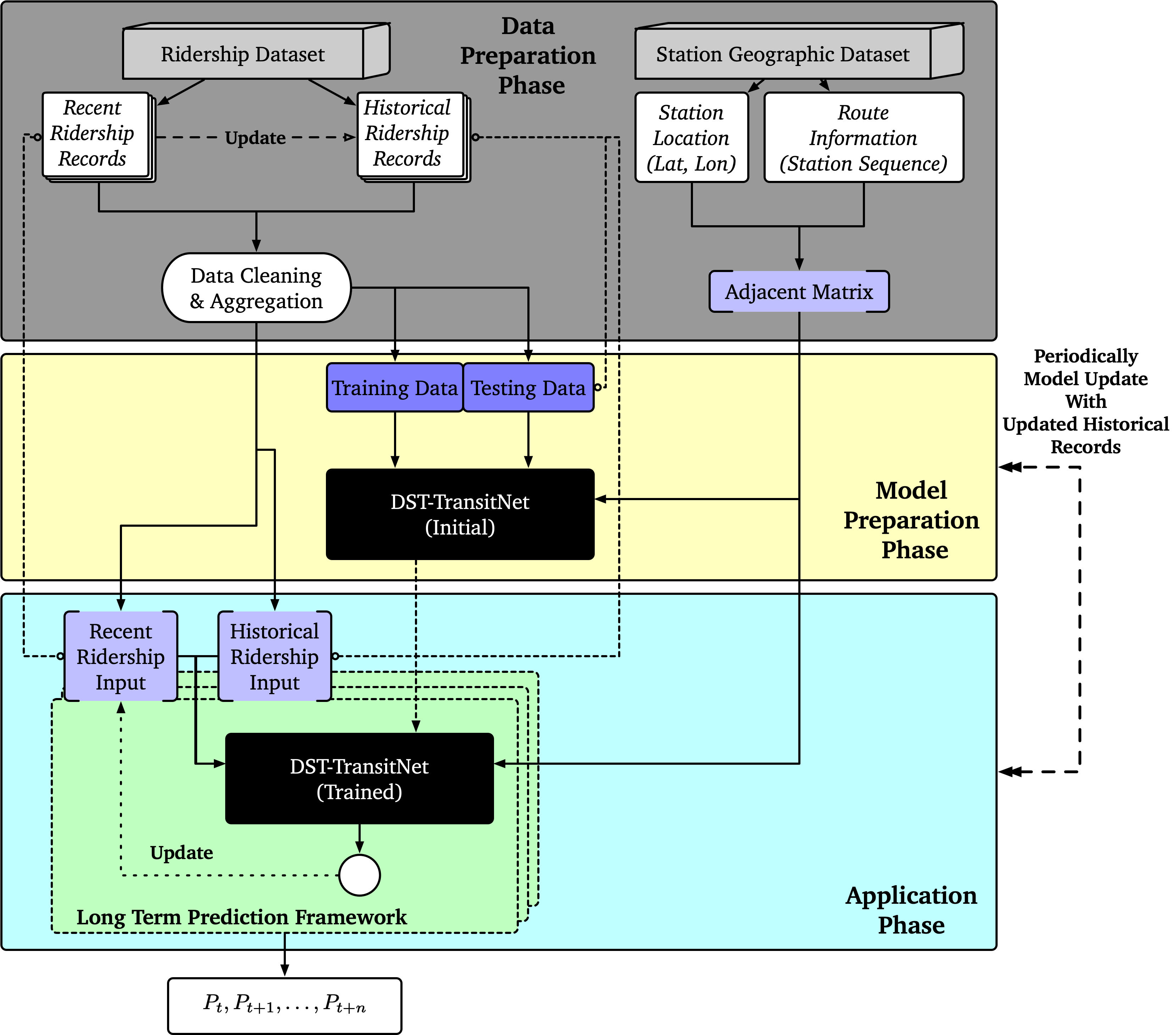}
    \caption{Pipeline for Real-World Deployment of DST-TransitNet}
    \label{fig:pipeline}
\end{figure}

As shown in Fig.~\ref{fig:pipeline}, the implementation process of DST-TransitNet can be divided into three main phases: Data Preparation, Model Preparation, and Application.

\textbf{Data Preparation Phase}:
\begin{itemize}
    \item \textbf{Data Collection}: Gather recent and historical ridership records, along with station geographic data, including station locations (latitude, longitude) and route information. The ridership dataset is dynamically updated with new records, whereas the station geographic dataset is relatively static and updated as needed.
    \item \textbf{Data Cleaning and Aggregation}: Clean and aggregate the collected data to prepare it for training and testing, including removing duplicates and aggregating data within the same time slots. Create an adjacency matrix using the station geographic dataset to represent the network structure.
\end{itemize}

\textbf{Model Preparation Phase}:
\begin{itemize}
    \item \textbf{Training Data and Testing Data}: Split the cleaned data into training and testing datasets derived from the historical ridership records.
    \item \textbf{Model Training}: Train the initial DST-TransitNet model using the training data. The model learns patterns from the historical data.
    \item \textbf{Model Validation}: Test the trained model on the testing data to validate its performance.
\end{itemize}

\textbf{Application Phase}:
\begin{itemize}
    \item \textbf{Input Data Requirements}: The deployed DST-TransitNet model requires recent ridership input collected from times $t-n$ to $t$ and historical ridership input collected from the previous day, covering times $t-n+1$ to $t+1$. This input data is used to predict future ridership at times $t+1$ for all stations across the system, where $n$ is the input length for the model.
    \item \textbf{Long Term Prediction Framework}: Within the long-term prediction framework, the model updates its recent ridership using its prediction results for the next time interval. Historical ridership input from the previous day, aligned with the prediction target time interval (e.g., $t+2$, $t+3$), is already in the dataset. Thus, for long-term predictions, DST-TransitNet requires only a single input from the ridership dataset to make multiple time predictions.
    \item \textbf{Continuous Learning or Periodic Fine-Tuning}: Periodically update (fine-tune) the model with new data to adapt to changing patterns and maintain accuracy. Fine-tuning reduces training costs and keeps the model aware of recent system changes.
\end{itemize}

This structured pipeline facilitates the effective implementation of DST-TransitNet, ensuring accurate and reliable ridership predictions for transit agencies.

\section{Experiments and Result Analysis}\label{sec:exp}
In this section, we detail the experiments conducted to evaluate the performance of the proposed DST-TransitNet model using real-world passenger count data from the BRT system of Bogota, Colombia. We outline the datasets used, describe the experimental setup, and present the evaluation metrics employed. Following this, we provide a comprehensive analysis of the results, highlighting the model's efficiency in predicting short-term station transit ridership across the network.

\subsection{Datasets}

\begin{figure}[!htbp]
    \centering
    \begin{subfigure}[t]{0.48\linewidth}
        \centering
        \includegraphics[width=\linewidth]{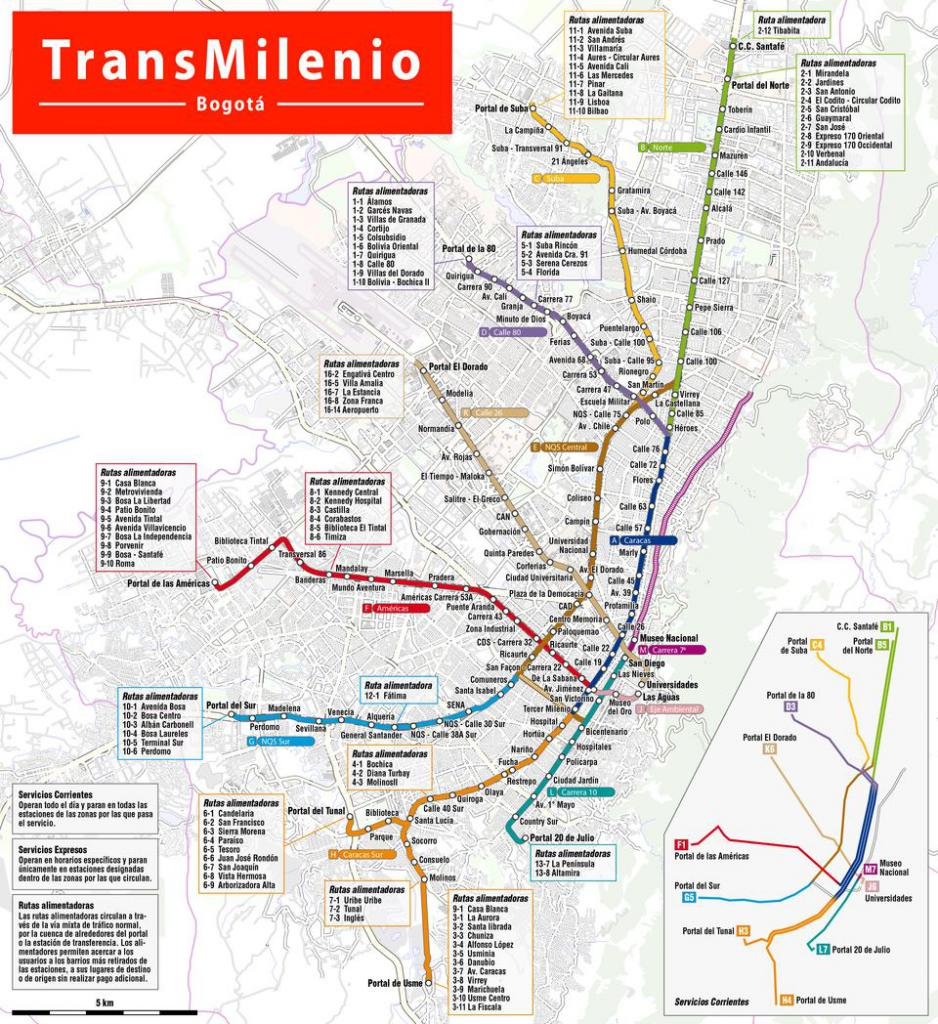}
        \caption{Official BRT System Map of Bogotá, Colombia~\footnote{https://development.asia/case-study/efficient-city-transport-those-who-do-not-own-cars}}
        \label{fig:transit_map}
    \end{subfigure}
    \hfill
    \begin{subfigure}[t]{0.48\linewidth}
        \centering
        \includegraphics[width=0.8\linewidth]{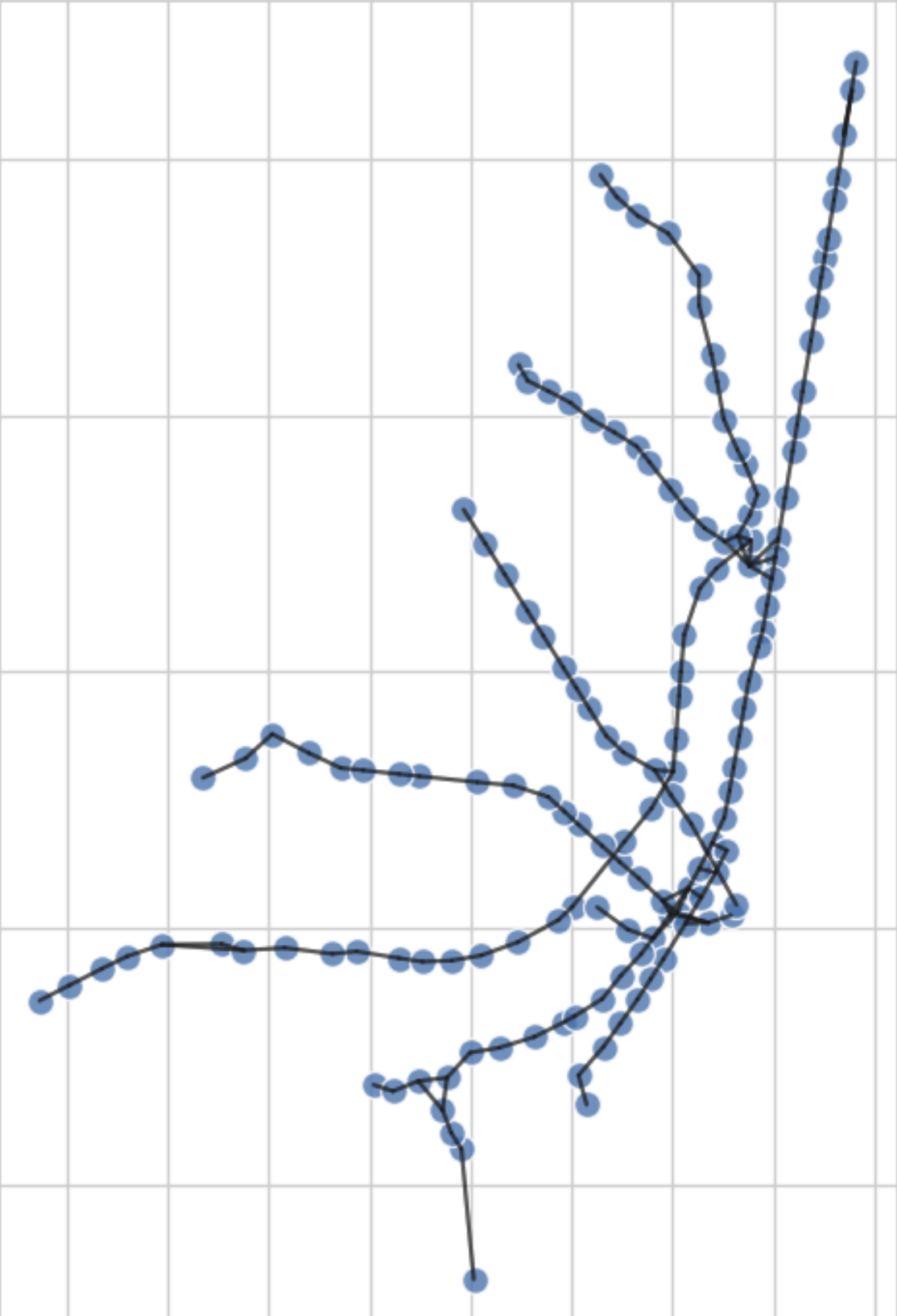}
        \caption{Geospatial Representation of BRT Stations and Connections in Bogotá, Colombia}
        \label{fig:station_map}
    \end{subfigure}
    \caption{Visual Comparison of BRT System Maps: Official vs. Geospatial Representation}
    \label{fig:comparison_maps}
\end{figure}

The dataset used for the experiments consists of passenger count records, specifically the number of boardings, collected from the TransMilenio Bus Rapid Transit (BRT) system in Bogotá, Colombia, as shown in Fig.~\ref{fig:comparison_maps}. Fig.~\ref{fig:transit_map} presents the official BRT system map, and Fig.~\ref{fig:station_map} provides a geospatial visualization of the BRT network, where nodes represent BRT stations and edges represent BRT routes between stations. The TransMilenio system comprises 12 BRT lines, serving an average of 2.3 million passengers daily and more than 250,000 passengers per hour during peak hours (pre-COVID-19), making it the busiest BRT system in the world~\cite{lemoine2016transmilenio}.
The passenger counts data is collected from 147 BRT stations over five years (from August 2015 to May 2021). In addition to the passenger counts information, the dataset also contains geometric information for each BRT station and connection information between stations.

\begin{figure}[!htbp]
    \centering
    \includegraphics[width=1\linewidth]{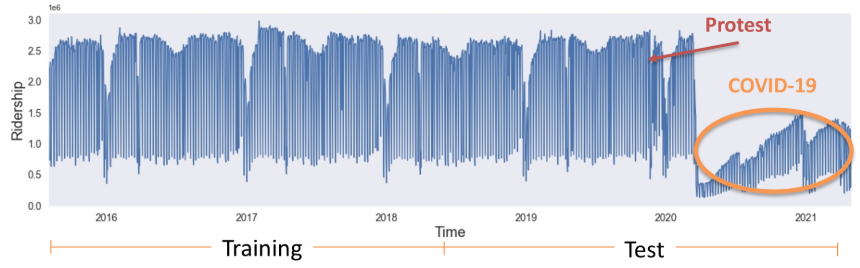}
    \caption{BRT Daily Aggregated Ridership From August 2015 to May 2021. Training Period: August 2015 to July 2018.
Test Period: August 2018 to May 2021. Protest: November and December 2019. COVID-19: March 2020 to May 2021. (Credit to~\cite{caicedo2023public}; Unit on the y-axis is \(10^6\))}
    \label{fig:ridership_sum}
\end{figure}

\begin{figure}[!htbp]
    \centering
    \includegraphics[width=1\linewidth]{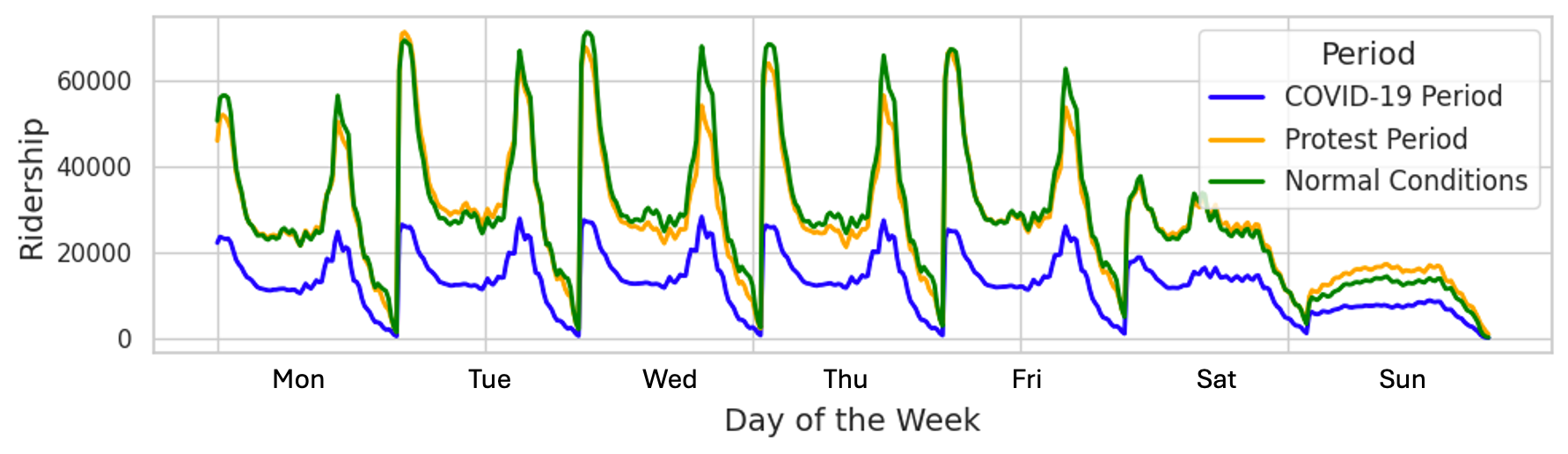}
    \caption{Weekly System Ridership for Different Periods}
    \label{fig:ridership_week}
\end{figure}

As introduced in~\cite{caicedo2023public}, where the data set was used for the prediction of daily ridership, the data set contains three different types of dynamics: the normal period (pre-COVID-19 period) from August 2015 to November 2019; the COVID-19 period, starting in March 2020 and continuing until the end of the dataset; and the protest period, which took place in November and December 2019, as shown in Fig.~\ref{fig:ridership_sum}. The dataset reveals a dramatic drop in ridership after the COVID-19 outbreak and varying ridership during the protest period. Fig.~\ref{fig:ridership_week} illustrates the average system ridership on a weekly basis across different periods in Bogota's BRT system.

For comparison purposes, we maintain the same dataset setting as in~\cite{caicedo2023public}, using ridership data before July 2018 for training. The test dataset includes ridership information from the normal period, protest period, and COVID-19 period. A key difference in our experiment is that we use a \textbf{15-minute} prediction interval for short-term ridership prediction, which includes more fluctuations and non-seasonal features, increasing the prediction difficulty. This interval is more useful for the management of the transit system, providing more flexibility to the management team by offering in-time and rapid system circumstance forecasting.

We compare our model with traditional ML methods such as FFNN and LSTM. We also include state-of-the-art time-series DL prediction models, iTransformer~\cite{liu2023itransformer} and DLinear~\cite{zeng2023transformers}. The iTransformer is based on the popular DL structure Transformer, tailored for time-series prediction. DLinear shares the same time decomposition layer as DST-TransitNet but only uses FFNN for feature extraction and prediction. Notably, except for our proposed models and DLinear, the rest of the tested models train and make predictions on a stop-by-stop basis, instead of training on all station information and making predictions at once.

The following are the basic experiment settings:
\begin{itemize}
    \item Number of target stops: 147
    \item Number of edges: 320
    \item Data aggregation interval: 15 minutes
    \item Model input length for recent ridership record: 20 (15 minutes each)
    \item Model input length for historical ridership record: 20 (15 minutes each, with the last record at the same time of the day and the same day of the week from the previous week)
    \item Prediction time interval: 15 minutes
    \item Test dataset sizes:
    \begin{itemize}
        \item Normal Period: (11,750, 147)
        \item COVID Period: (19,317, 147);
        \item Protest Period: (2,820, 147).
    \end{itemize}
\end{itemize}

\subsection{Evaluation Metrics}
To evaluate the model's performance, we utilize two key evaluation metrics: the R-squared value (\(R^2\)) and the Mean Arctangent Absolute Percentage Error (MAAPE)~\cite{kim2016new}. 

The \(R^2\) value, calculated as follows:
\begin{equation}
    R^2 = 1 - \frac{\sum_{i=1}^{n} (y_i - \hat{y}_i)^2}{\sum_{i=1}^{n} (y_i - \bar{y})^2},
    \label{equ:r2}
\end{equation}
is a well-known statistical measure that indicates the proportion of variance in the dependent variable that is predictable from the independent variables. It provides a clear indication of how well the model's predictions align with the actual observed values, with a value closer to 1 signifying a better fit.

On the other hand, MAAPE, calculated as follows:
\begin{equation}
    \text{MAAPE} = \frac{1}{n} \sum_{i=1}^{n} \arctan\left(\left|\frac{y_i - \hat{y}_i}{y_i}\right|\right),
\end{equation}
is a relatively new metric introduced to address some limitations of traditional error metrics like Mean Absolute Percentage Error (MAPE). Unlike MAPE, which can be disproportionately affected by small denominators leading to large error values, MAAPE provides a more balanced assessment by incorporating the arctangent function. This enhances the robustness of the error measure, making it useful for evaluating models in the presence of noisy data.

Using both \(R^2\) and MAAPE, we ensure a comprehensive evaluation of the performance of the model, capturing both the goodness of fit and the robustness against outliers and variability in the data. This dual-metric approach allows for a more nuanced understanding of the model's predictive capabilities and reliability across different scenarios.

\subsection{Results and Analysis}
In this section, we present a comprehensive analysis of the performance of the proposed DST-TransitNet models, along with the several aforementioned benchmark models, across different dataset periods for short- and long-term prediction tasks.
\subsubsection{Model Performance Overview}

\begin{figure}[!htbp]
    \centering
    \begin{subfigure}[t]{1\linewidth}
        \centering
        \includegraphics[width=\linewidth]{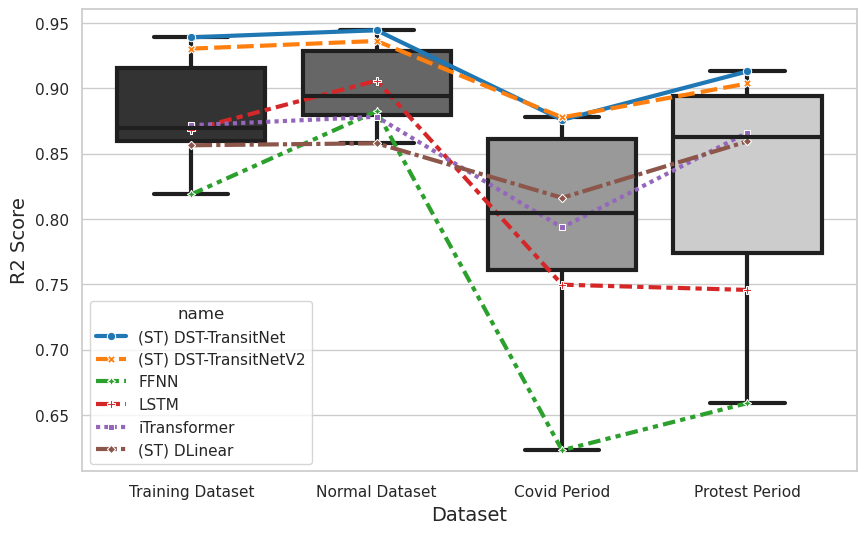}
        \caption{Dataset-level R$^2$ Score Distribution}
        \label{fig:dataset_r2_distribution}
    \end{subfigure}
    \hfill
    \begin{subfigure}[t]{1\linewidth}
        \centering
        \includegraphics[width=\linewidth]{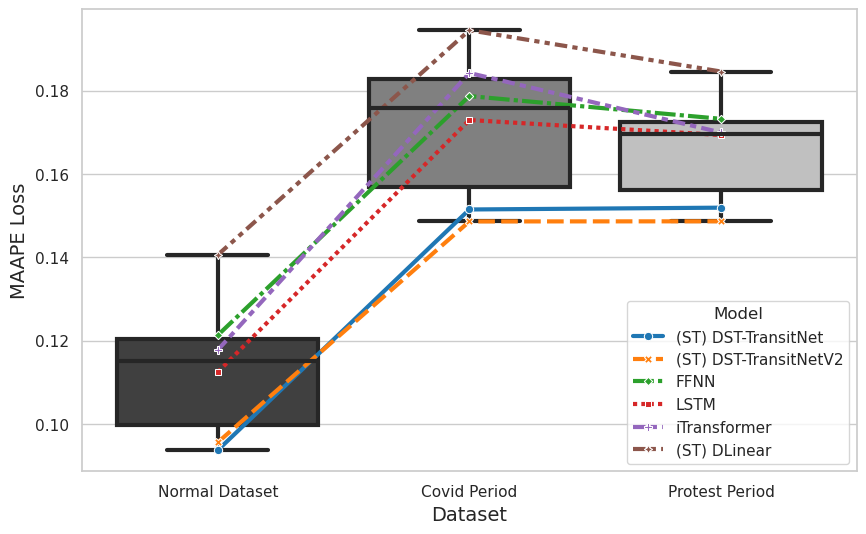}
        \caption{Dataset-level MAAPE Distribution}
        \label{fig:dataset_maape_distribution}
    \end{subfigure}
    \caption{Performance comparison of different models across various dataset periods. The "(ST)" appearing before some model names indicates the model makes predictions for all stops in the transit network simultaneously.}
    \label{fig:performance_comparison}
\end{figure}

Fig.~\ref{fig:performance_comparison} and Table~\ref{tab:maape_r2_scores} illustrate the $R^2$ score and MAAPE distribution for each model across the training dataset, normal period, COVID-19 period, and protest period. The results demonstrate that the two proposed models consistently achieve the best performance across all testing periods, indicating their superior predictive capabilities. This highlights the strong ability of the proposed models to capture underlying patterns and generalize well to unseen stable data.
\begin{table*}[!htbp]
    \centering
    \renewcommand{\arraystretch}{1.2} % Adjust row height
    \setlength{\tabcolsep}{0.5pt} % Adjust column width
    \begin{tabular}{|c|c|c|c|c|c|c|c|c|c|}
        \hline
        \textbf{Model}        & \textbf{Dataset}                        & \textbf{MAAPE} & \textbf{R2} & \textbf{Dataset}                       & \textbf{MAAPE} & \textbf{R2} & \textbf{Dataset}                  & \textbf{MAAPE} & \textbf{R2} \\ \hline
        FFNN                  & \multirow{6}{*}{\textbf{Normal}} & 0.1213              & 0.8825            & \multirow{6}{*}{\textbf{Covid}} & 0.1787              & 0.6230            & \multirow{6}{*}{\textbf{Protest}} & 0.1732              & 0.6591            \\ \cline{1-1} \cline{3-4} \cline{6-7} \cline{9-10} 
        LSTM                  &                                         & 0.1124              & 0.9059            &                                        & 0.1729              & 0.7497            &                                   & 0.1693              & 0.7457            \\ \cline{1-1} \cline{3-4} \cline{6-7} \cline{9-10} 
        iTransformer          &                                         & 0.1177              & 0.8782            &                                        & 0.1842              & 0.7935            &                                   & 0.1699              & 0.8662            \\ \cline{1-1} \cline{3-4} \cline{6-7} \cline{9-10} 
        (ST) DLinear          &                                         & 0.1406              & 0.8579            &                                        & 0.1945              & 0.8162            &                                   & 0.1846              & 0.8594            \\ \cline{1-1} \cline{3-4} \cline{6-7} \cline{9-10} 
        (ST) DST-TransitNet   &                                         & \textbf{0.0937}     & \textbf{0.9444}   &                                        & \textbf{0.1515}     & \textbf{0.8759}   &                                   & \textbf{0.1519}     & \textbf{0.9130}   \\ \cline{1-1} \cline{3-4} \cline{6-7} \cline{9-10} 
        (ST) DST-TransitNetV2 &                                         & \textbf{0.0955}     & \textbf{0.9362}   &                                        & \textbf{0.1486}     & \textbf{0.8777}   &                                   & \textbf{0.1486}     & \textbf{0.9037}   \\ \hline
    \end{tabular}
    \caption{MAAPE Loss and R2 Score for Different Models Across Normal, COVID-19, and Protest Periods.}
    \label{tab:maape_r2_scores}
\end{table*}

\begin{figure}[htbp!]
    \centering
    \begin{subfigure}[t]{1\linewidth}
        \centering
        \includegraphics[width=\linewidth]{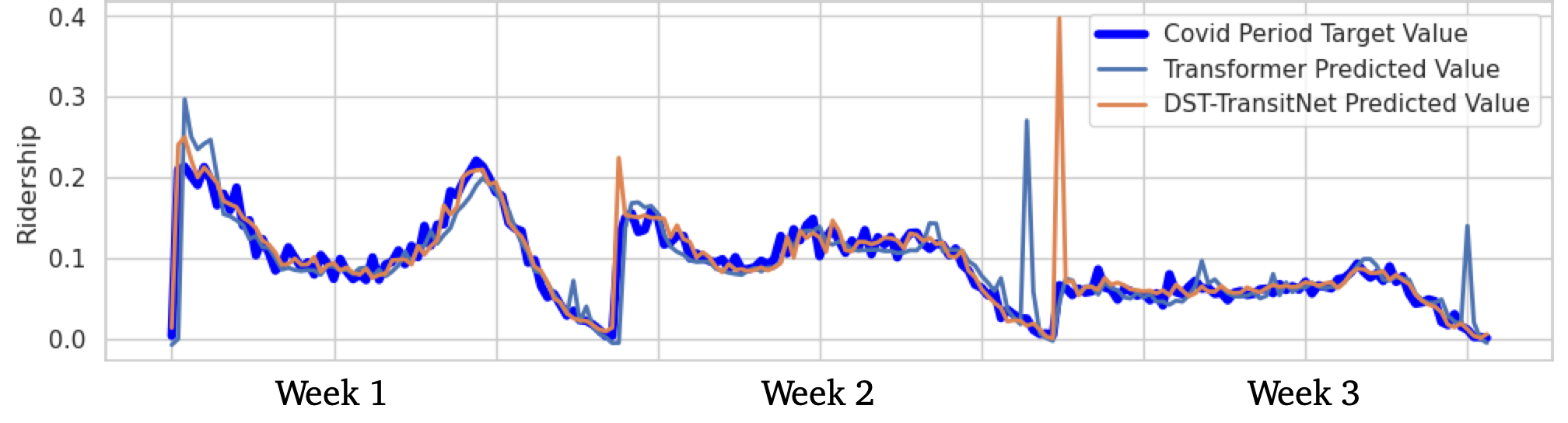}
        \caption{Covid Period}
        \label{fig:covid_comparison}
    \end{subfigure}
    \hfill
    \begin{subfigure}[t]{1\linewidth}
        \centering
        \includegraphics[width=\linewidth]{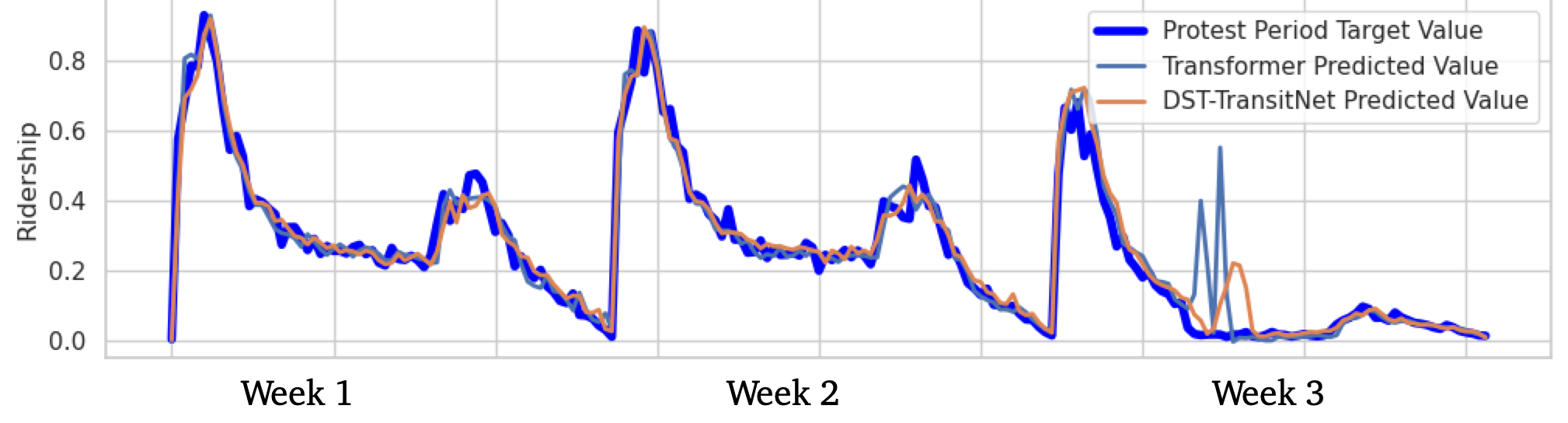}
        \caption{Protest Period}
        \label{fig:protest_comparison}
    \end{subfigure}
    
    \caption{Comparison of the prediction results from the Transformer model and our proposed DST-TransitNet model on the (scaled) ridership for one station (id: 2000) in three consecutive weeks. The predicted values are compared with the target values.}
    \label{fig:comparison_predictions}
\end{figure}

As shown in Fig.~\ref{fig:comparison_predictions}, the prediction results from the Transformer model and our proposed DST-TransitNet model on the scaled ridership for station 2000 are compared with the target values over three selected consecutive weeks during the Covid period and Protest period. During both periods, both models perform well in the middle of the day. However, DST-TransitNet demonstrates superior performance during peak hours, showing less deviation from the target values. Furthermore, when the system encounters unforeseen disruptions, such as the transit workers' protest starting in week 3 of the Protest period, DST-TransitNet adapts more quickly and smoothly to the abnormal situation, providing more accurate predictions. Notably, at the start of week 3 during the Covid period, both models exhibit significant prediction errors. This is because the models expect ridership to shift from low to high levels during the morning peak. However, due to Covid-19, there was no morning peak in week 3, leading to relatively large prediction errors.

Notably, the performance all models on the normal period testing dataset is slightly better than on the training dataset, which means all models are well trained and do not exhibit an overfitting problem. However, during periods of external disruptions, such as the COVID-19 pandemic and national protests, the performance of all models declines, reflecting the increased difficulty of accurate prediction in highly volatile environments. Notably, DST-TransitNet and DST-TransitNetV2 maintain the best performance compared with other models, showcasing the resilience and adaptability of the structure, which is important for real-world applications, where models must handle unforeseen events. The performance of the models underscores the advantages of spatio-temporal models in maintaining predictive accuracy across diverse conditions.

Furthermore, when comparing the performance of FFNN and DLinear, we observe that DLinear not only exhibits overall better performance across all prediction scenarios but also shows a smaller performance drop when external disturbances are introduced to the dataset. This suggests that incorporating the time-series decomposition framework in model construction not only enhances accuracy and interpretability but also improves the model's stability in changing prediction environments.

\begin{figure}[!htbp]
    \centering
    \begin{subfigure}[t]{1\linewidth}
        \centering
        \includegraphics[width=\linewidth]{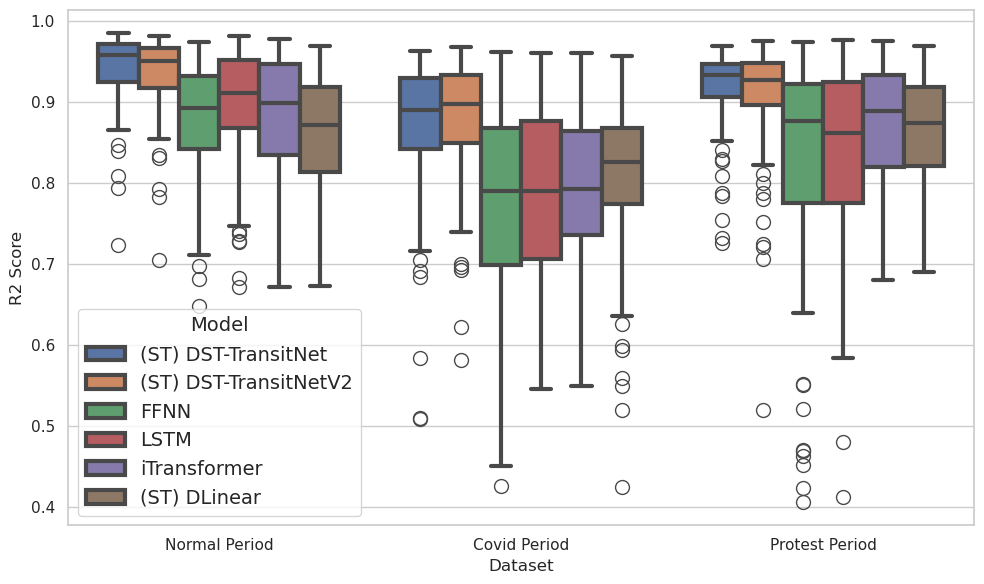}
        \caption{Network-wise Station-level R$^2$ Score Distribution}
        \label{fig:r2_distribution}
    \end{subfigure}
    \hfill
    \begin{subfigure}[t]{1\linewidth}
        \centering
        \includegraphics[width=\linewidth]{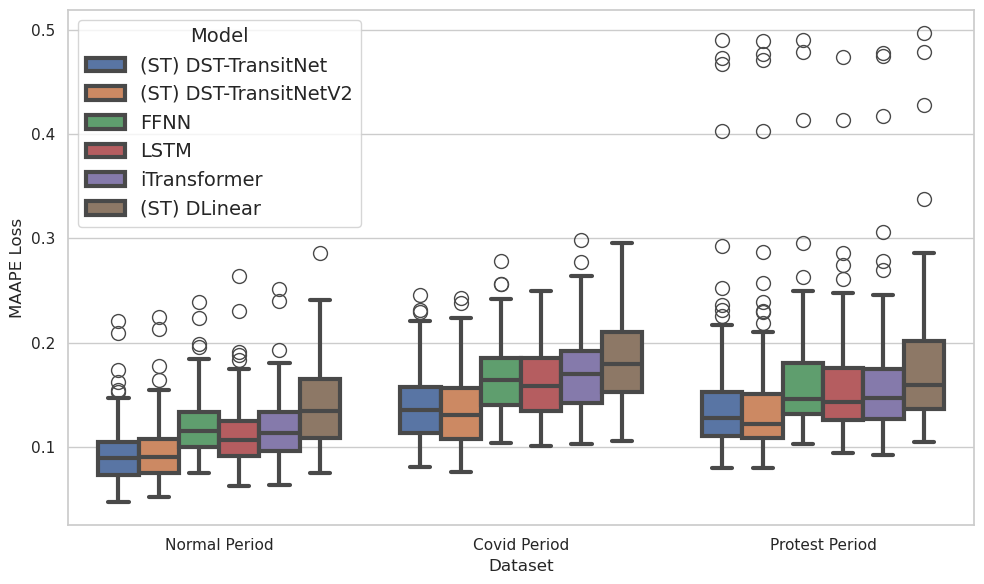}
        \caption{Network-wise Station-level MAAPE Distribution}
        \label{fig:maape_distribution}
    \end{subfigure}
    \caption{Criteria Distribution Among Stations Across Different Models and Prediction Scenarios.}
    \label{fig:performance_comparison_2}
\end{figure}

The Network-wide Station-level MAAPE loss and $R^2$ distribution (Fig.~\ref{fig:performance_comparison_2}) indicate that the new models are consistently more accurate and robust across all periods, with lower MAAPE values/higher $R^2$ score and narrower interquartile ranges (IQRs). This suggests that the proposed models are better at handling both stable and rapidly changing conditions for all stations in the network.

\begin{figure}[!htbp]
    \centering
    \includegraphics[width=1\linewidth]{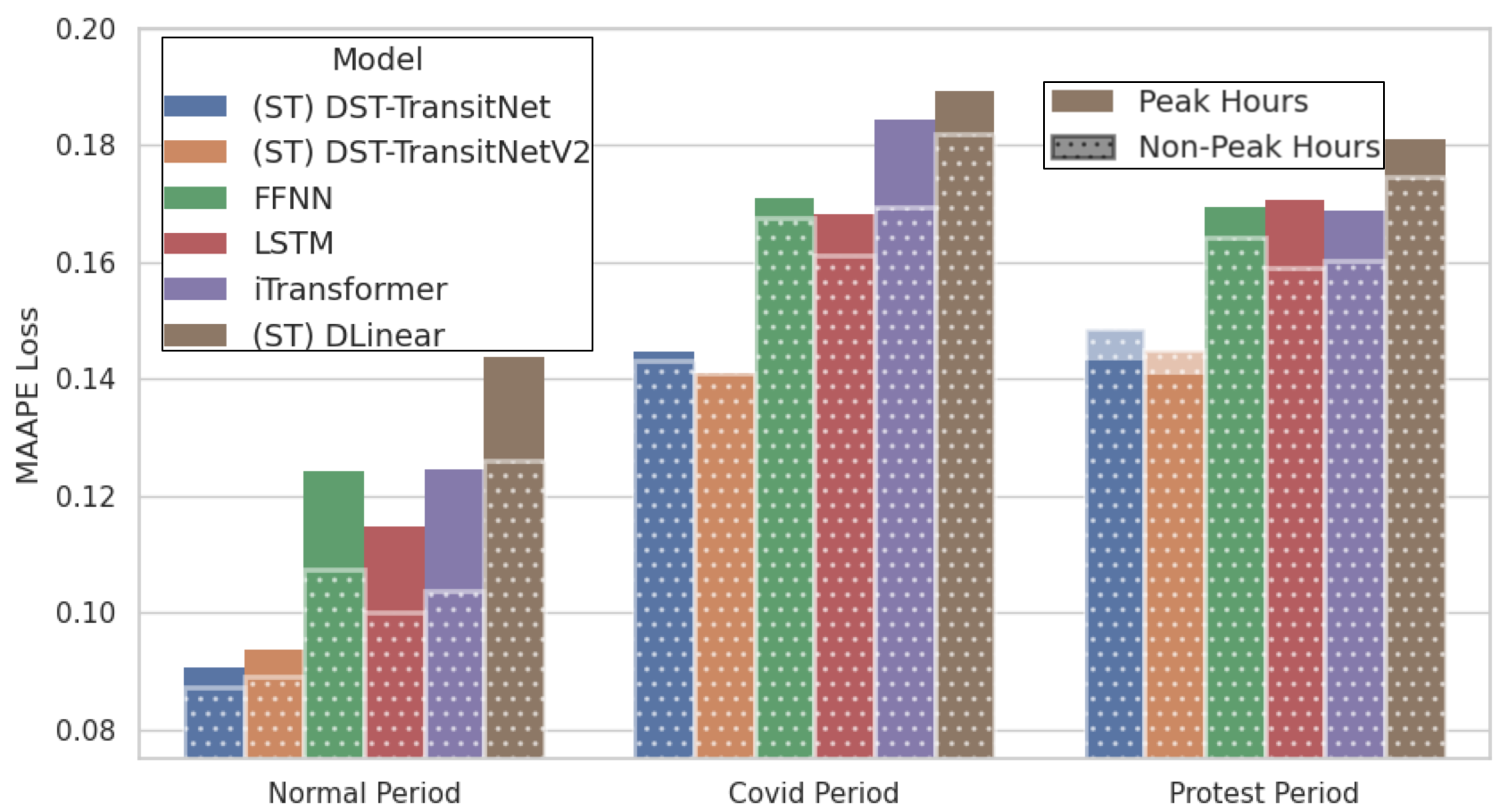}
    \caption{MAAPE Across Different Models and Periods During Peak and Non-Peak Hours}
    \label{fig:performance_comparison_3}
\end{figure}

Figure~\ref{fig:performance_comparison_3} highlights the variation in model performance during peak and non-peak hours. Peak hours are defined as two time slices: from 6 AM to 10 AM, and from 5 PM to 9 PM. It is observed that all models perform worse during peak hours, indicated by higher MAAPE loss values, compared to non-peak hours. This performance degradation is primarily due to the higher variability in transit ridership during peak hours, caused by increased passenger volume and congestion, which introduces more noise into the dataset. Despite the more challenging prediction environment, DST-TransitNet models markedly outperform other models during both peak and non-peak hours. 

\begin{figure}[htbp!]
    \centering
    \begin{subfigure}{0.8\linewidth}
        \includegraphics[width=\linewidth]{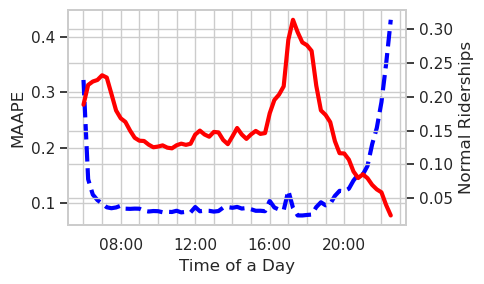}
        \caption{Ridership and MAAPE loss for Normal ridership period.}
        \label{fig:normal_ridership}
    \end{subfigure}
    \hfill
    \begin{subfigure}{0.8\linewidth}
        \includegraphics[width=\linewidth]{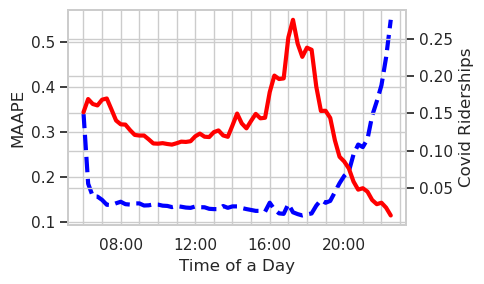}
        \caption{Ridership and MAAPE loss for COVID-19 period.}
        \label{fig:covid_ridership}
    \end{subfigure}
    \hfill
    \begin{subfigure}{0.8\linewidth}
        \includegraphics[width=\linewidth]{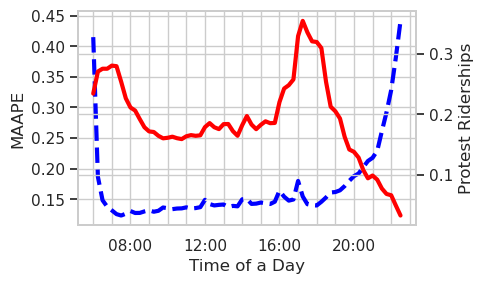}
        \caption{Ridership and MAAPE loss for Protest ridership.}
        \label{fig:protest_ridership}
    \end{subfigure}
    \caption{MAAPE and target ridership values for different periods. Each subplot corresponds to a different period: normal, COVID-19, and protest. The MAAPE is calculated and plotted alongside the target ridership values to analyze prediction accuracy across different times of the day. The dashed blue line indicates the MAAPE values, and the solid red line represents the target ridership values. The x-axis spans from 6am to 11:59pm with 15-minute intervals.}
    \label{fig:ridership_analysis}
\end{figure}

Meanwhile, we also notice that the performance gap between peak and non-peak hours becomes smaller as the prediction task complexity increases.
To further elucidate the model's performance variation during non-peak and peak hours, we present Fig~\ref{fig:ridership_analysis}. This figure clearly illustrates the MAAPE loss for the FFNN model across different times of the day. Notably, the MAAPE loss increases as the prediction target approaches both the onset and the conclusion of peak hours. During all periods, higher MAAPE losses are observed at the beginning of the morning peak and the end of the evening peak. The MAAPE continues to rise until the end of the day, contributing to higher non-peak hour values.

During the COVID-19 and protest periods, higher non-peak hour MAAPE losses occur closer to the morning peak during the COVID-19 period. In contrast, during the protest period, elevated MAAPE losses begin in the morning and persist throughout the day. These differences can be attributed to the time-series changes unique to each period.

\begin{figure}[!htbp]
    \centering
    \includegraphics[width=1\linewidth]{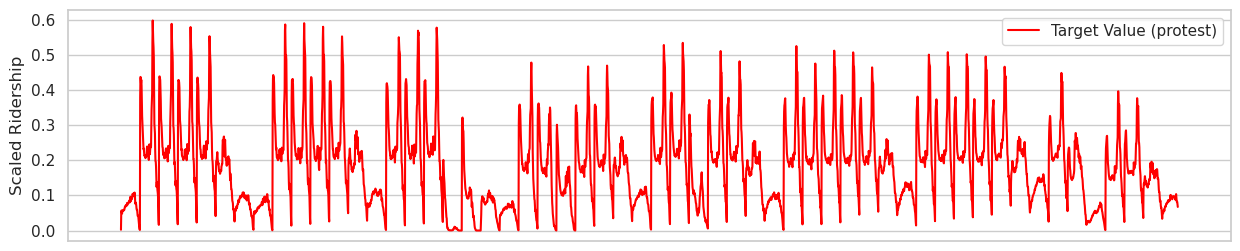}
    \caption{Scaled Ridership During Protest Period}
    \label{fig:prot_ride}
\end{figure}

For the COVID-19 period, as shown in Fig.~\ref{fig:ridership_week}, the overall ridership drops, resulting in smoother transitions between peak and non-peak hours. Consequently, the farther the prediction is from the morning peak, the lesser the impact, leading to relatively lower MAAPE losses. However, during the protest period, illustrated in Fig.~\ref{fig:prot_ride}, irregular ridership patterns due to system shutdowns and irregular running times cause more variability. In this scenario, the distinction between peak and non-peak hours has a lesser impact on ridership changes compared to the system's operational irregularities. As the day progresses, these irregularities accumulate in the model's inputs, leading to increasing MAAPE losses.

\begin{figure*}[!htbp]
    \centering
    \includegraphics[width=1\linewidth]{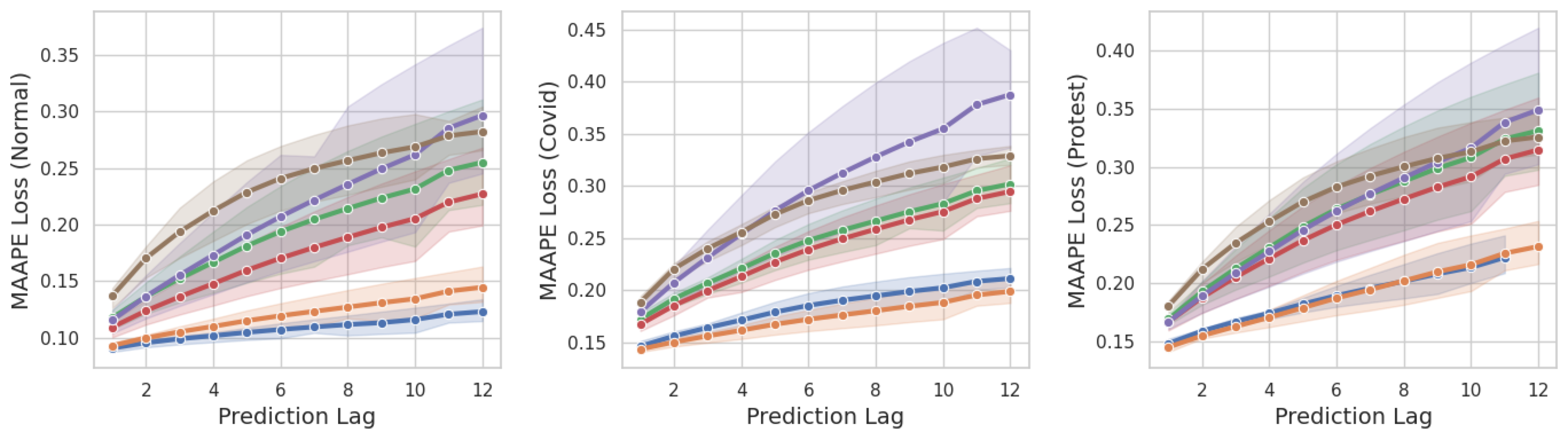}
    \caption{MAAPE Loss Across Different Models and Prediction Lags During Normal, COVID-19, and Protest Periods. The solid line in each plot represents the loss value of each model at the corresponding prediction lag, and the shaded area indicates the range between peak and non-peak hour MAAPE values.}
    \label{fig:mlncp}
\end{figure*}

Moving to the long-term prediction performance using the methods introduced earlier, Table~\ref{tab:maape_ratio} shows the increasing ratio of the MAAPE loss for each prediction model when the prediction interval increases from 1 time interval to 12 time intervals. In our experimental setup, this corresponds to making ridership predictions from the next 15 minutes to the next 3 hours. Figure~\ref{fig:mlncp} illustrates the MAAPE loss across different models for all prediction lags from 1 to 12.
As observed in the figure, all models exhibit a drop in performance as the prediction interval increases. As explained earlier, prediction errors accumulate in the model's input, deteriorating the final prediction results. However, our proposed models not only demonstrate the best performance across all prediction lags, maintaining an increasing lead over other models, but also exhibit the most stable prediction performance as the prediction lag increases. While it is common for other models to show a larger performance gap between peak and non-peak hours with increasing prediction lag, our proposed models maintain this distance within a much smaller and steadier range. This demonstrates the remarkable stability and robustness of our proposed models in various prediction scenarios, making them more suitable for real-world applications with broader application scenarios.
\begin{table}[!htbp]
    \centering
    \renewcommand{\arraystretch}{1} % Adjust row height
    \setlength{\tabcolsep}{10pt} % Adjust column width
    \begin{tabular}{|l|l|c|}
        \hline
        \textbf{Model} & \textbf{Dataset} & \textbf{MAAPE Ratio} \\ \hline
        \multirow{3}{*}{(ST) DST-TransitNet}   
            & \textbf{Normal Period}  & \textbf{1.265272} \\ \cline{2-3} 
            & Covid Period   & 1.364819 \\ \cline{2-3} 
            & \textbf{Protest Period} & \textbf{1.423226} \\ \hline
        \multirow{3}{*}{(ST) DST-TransitNetV2} 
            & Normal Period  & 1.424043 \\ \cline{2-3} 
            & \textbf{Covid Period}   & \textbf{1.293841} \\ \cline{2-3} 
            & Protest Period & 1.472718 \\ \hline
        \multirow{3}{*}{LSTM}                  
            & Normal Period  & 1.846863 \\ \cline{2-3} 
            & Covid Period   & 1.630786 \\ \cline{2-3} 
            & Protest Period & 1.729708 \\ \hline
        \multirow{3}{*}{FFNN}                  
            & Normal Period  & 1.906202 \\ \cline{2-3} 
            & Covid Period   & 1.624358 \\ \cline{2-3} 
            & Protest Period & 1.782810 \\ \hline
        \multirow{3}{*}{(ST) DLinear}          
            & Normal Period  & 1.940223 \\ \cline{2-3} 
            & Covid Period   & 1.674803 \\ \cline{2-3} 
            & Protest Period & 1.717645 \\ \hline
        \multirow{3}{*}{iTransformer}          
            & Normal Period  & 2.214277 \\ \cline{2-3} 
            & Covid Period   & 1.917513 \\ \cline{2-3} 
            & Protest Period & 1.857993 \\ \hline
    \end{tabular}
    \caption{Comparison of MAAPE Ratios for Different Prediction Lags (12 vs. 1) Across Various Models and Datasets. The MAAPE ratio reflects the relative prediction error increase from short-term (lag 1) to longer-term (lag 12) predictions, highlighting model performance stability over extended periods.}
    \label{tab:maape_ratio}
\end{table}

Finally, we present the training times and model sizes for deploying the ML and DL models for the short-term ridership prediction task, as shown in Table~\ref{tab:training_time}. The training time represents the total time required to train models for all stations. LSTM, FFNN, and iTransformer perform one-to-one predictions, meaning each model only takes information from one stop to make a prediction for that specific stop. Consequently, 147 different models are required to make predictions for the entire system.
\begin{table*}[!htbp]
    \centering
    \renewcommand{\arraystretch}{1.2} % Adjust row height
    \setlength{\tabcolsep}{4pt} % Adjust column width
    \begin{tabular}{c|cc|cc}
        \toprule
        \textbf{Model} & \textbf{Training Time (min)} & \textbf{Rel. Time (\%)} & \textbf{Size (MB)} & \textbf{Rel. Size (\%)} \\ \midrule
        LSTM                  & 723.8 & 100 (Baseline) & 5.75  & 100 (Baseline) \\ 
        FFNN                  & 355.6 & 49.1  & 1.83  & 31.8  \\ 
        iTransformer          & 292.5 & 40.4  & 3805  & 66173.9  \\ 
        (ST) DST-TransitNet   & 44.7  & 6.2   & 0.8   & 13.9  \\ 
        (ST) DLinear          & \textbf{36.0}  & \textbf{5.0}   & \textbf{0.3}   & \textbf{5.2}   \\ 
        (ST) DST-TransitNetV2 & \textbf{33.8}  & \textbf{4.7}   & \textbf{0.5}   & \textbf{8.7}   \\ \bottomrule
    \end{tabular}
    \caption{Training time and model size comparison of different models.}
    \label{tab:training_time}
\end{table*}

In contrast, due to the spatio-temporal structure of DST-TransitNets and DLinear, these models can perform many-to-many predictions, requiring only one model for the whole system. Although the training time per model may be longer for many-to-many models due to their complexity, they still save substantial resources as the size of the prediction target increases. Additionally, the model sizes vary significantly, with DST-TransitNets being more compact compared to the iTransformer, which has an exceptionally large model size.

\subsubsection{Analysis of Results from Temporal and Spatial Perspectives}

In the previous subsection, we discuss the performance of our model and other ML/DL models from a temporal perspective, while in this subsection, we further interpret the performance of the model from both temporal and spatial perspectives jointly. Analyzing the results through these lenses helps us understand the underlying reasons for the model's behavior and provides insights for future research and model design. By dissecting the temporal and spatial patterns, we can identify key factors that influence model accuracy and robustness, thereby guiding the development of more effective predictive models.

From a spatial perspective, we evaluate the model's performance across different geographic locations within the transit network. This involves analyzing the model performance, considering factors such as station connectivity, and geographic characteristics.

\begin{figure}[!htpb]
    \centering
    \includegraphics[width=1\linewidth]{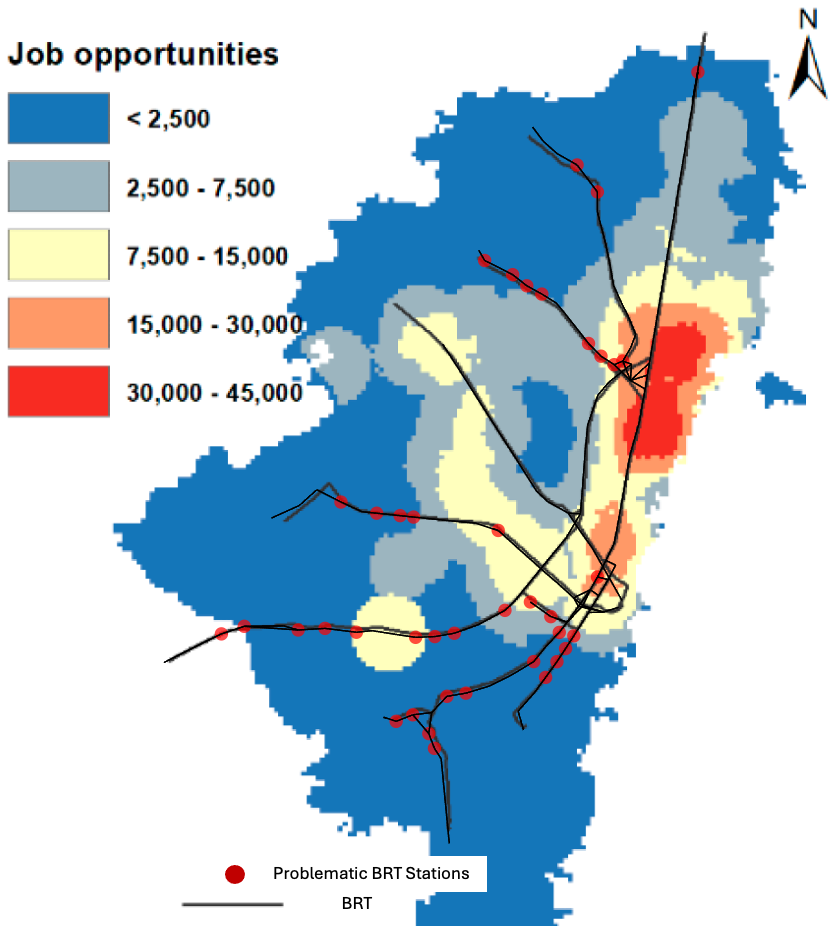}
    \caption{Geographical distribution of challenge BRT stations in relation to job opportunities. The map shows areas of varying job opportunities, with challenge stations marked in red.}
    \label{fig:geo_location_of_problematic_stations}
\end{figure}

\begin{figure}[htbp!]
    \centering
    \includegraphics[width=1\linewidth]{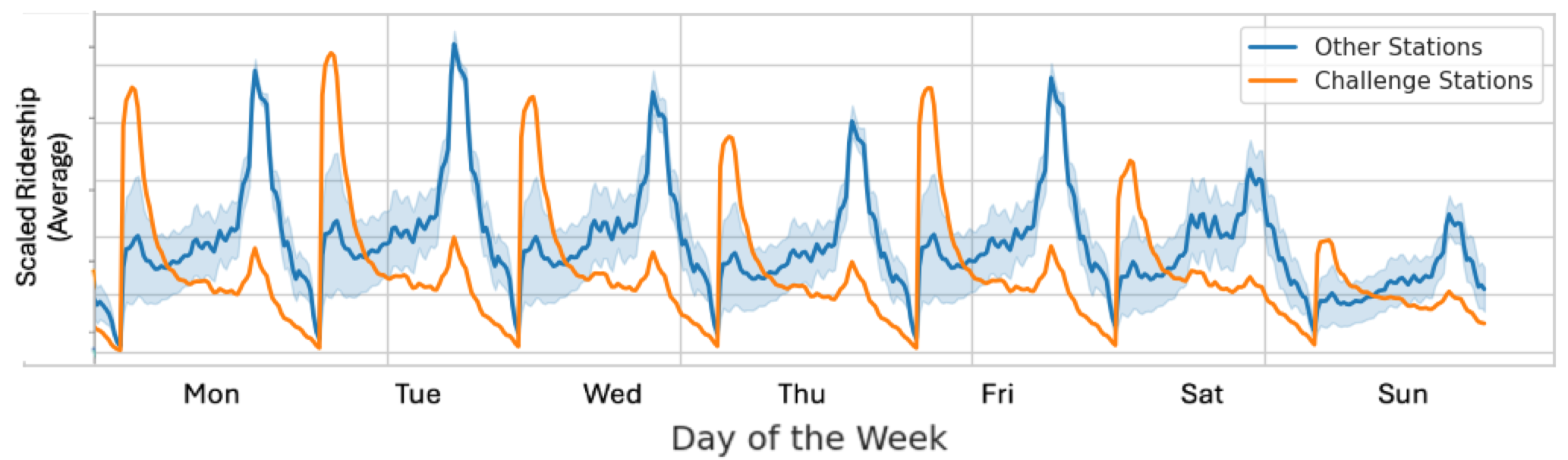}
    \caption{Comparison of the average scaled ridership between the challenge stations and the rest of the system as a whole over a week.}
    \label{fig:system_avg_prob_scal}
\end{figure}

To begin, we define stations with an \(R^2\) score (defined in Equ.~\ref{equ:r2}) lower than 80\% of all stations as "challenge stations."
Due to the lack of accessible census data for the City of Bogota, we only link the challenge stations' spatial information with the job density geo-information in Fig.~\ref{fig:geo_location_of_problematic_stations}. As we can see, job density decreases from the center of the city to the mid-ring of the city, and to the outskirts of the city. Generally, areas with higher job density are business centers, while lower density areas are typically residential or rural.
The challenge stations mainly lie in two types of areas: first, the edge of multi-type fields, and second, in residential areas close to the end of the BRT route. The possible reasons for the model's poor performance could be attributed to two main factors. First, for stations located on the edge, each target station might exhibit different ridership patterns from its neighboring stations and thus the association between these station might be weak. Second, for stations in residential or rural areas, the ridership patterns may significantly differ from the main portion of the dataset. As illustrated in Fig.~\ref{fig:system_avg_prob_scal}, which shows the average scaled ridership for all challenge stations in residential/rural areas and the rest stations on a weekly basis, the time series pattern of the challenge stations is notably different. Specifically, challenge stations exhibit only a single morning peak hour throughout the day. This discrepancy is reasonable, as it reflects people leaving home for work or school in the morning. As mentioned earlier, the dataset excludes ridership records from 12 AM to 6 AM, causing a sudden increase in ridership from 0 to a relatively high level at these stations during the morning peak, which makes prediction more challenging.

By understanding these spatial patterns, we can refine the model to account for location-specific factors. Segregating the stations into groups based on land use or other geographical information and training and testing them separately could be beneficial for improving overall accuracy and reliability.

\begin{figure}[!htbp]
    \centering
    \includegraphics[width=1\linewidth]{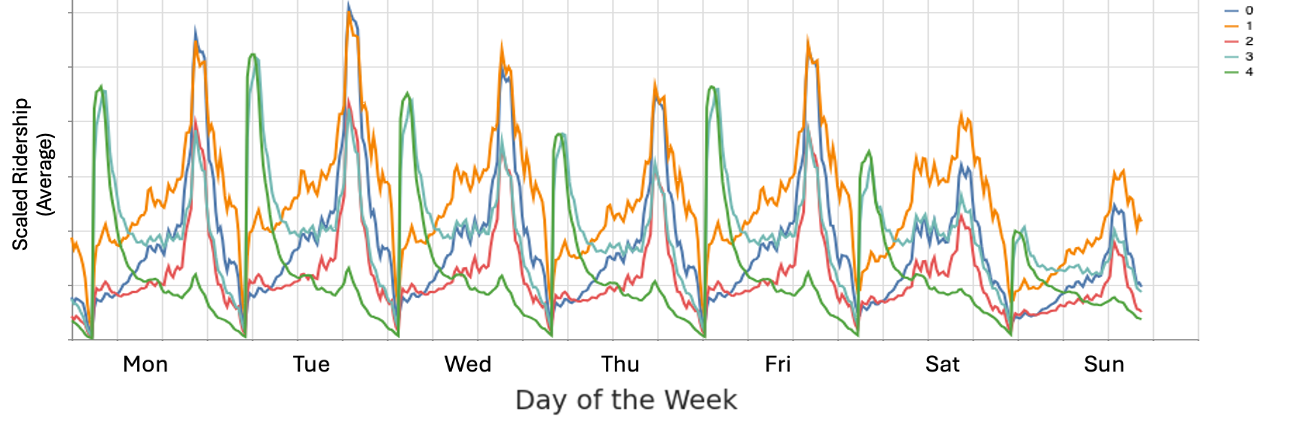}
    \caption{Average scaled ridership for different station groups over a week.}
    \label{fig:group_scaled_week}
\end{figure}

Lastly, we highlight the combination of both temporal and spatial analyses. This dual approach enables us to pinpoint specific temporal-spatial interactions that may affect model predictions. We used the KMeans clustering method to group all stations in the dataset into five groups based on the aggregated mean ridership on a weekly basis for each station. In Fig.~\ref{fig:group_scaled_week}, we show the average scaled ridership for each group over a week. It is clear that each group exhibits distinct temporal patterns.
For Groups 0, 1, and 2, there is only one evening peak during weekdays. However, Group 0's shape is sharper, indicating that the passenger count remains high for a shorter duration. In contrast, Group 1, while having a similar peak value to Group 0, maintains a higher level of passenger demand throughout the day. Group 2 also has one peak period throughout the day but with a smaller difference between peak and non-peak hours.
Group 3 shows a two-peak pattern similar to the system-level average, with distinct morning and evening peaks. Finally, Group 4 exhibits only a morning peak.

\begin{figure}[htbp!]
    \centering
    \includegraphics[width=1\linewidth]{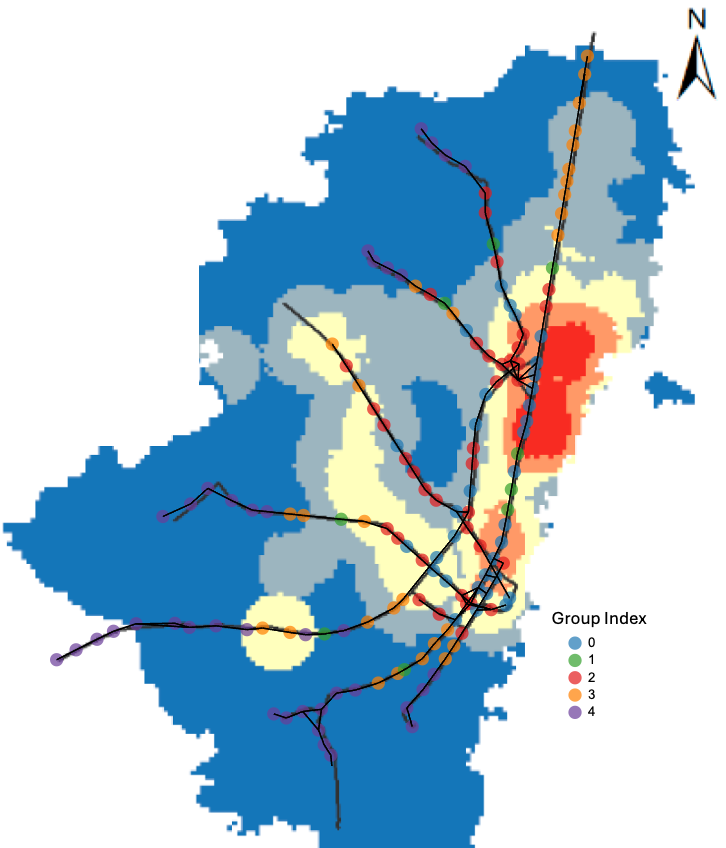}
    \caption{Geographical distribution of station groups overlaid on job density.}
    \label{fig:knn_dis}
\end{figure}

After mapping the station groups onto the geographical map, as shown in Fig.~\ref{fig:knn_dis}, we observed that even though the clustering algorithm only considered historical ridership information, the stations within the same group tend to share similar spatial features. These groups align with the job density distribution in the city. For instance, stations in Group 0 predominantly appear in the city center/business areas, while stations in Group 4 are mostly located in low job density areas. Additionally, Group 1 and Group 2 stations are primarily situated on the edges of different land use areas.

\begin{figure}[htbp!]
    \centering
    \includegraphics[width=1\linewidth]{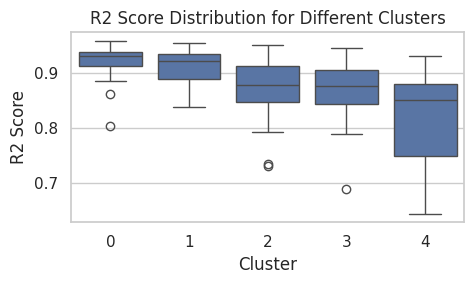}
    \caption{R2 score distribution for different station clusters. The performance declines from Group 0 to Group 4, with Group 4 encompassing most of the challenge stations.}
    \label{fig:r2_group}
\end{figure}

Finally, we present the model performance across different groups in Fig.~\ref{fig:r2_group}. It is evident that the prediction performance is highly related to the clustered groups. The performance decreases progressively from Group 0 to Group 4. Notably, Group 4 includes almost all the challenge stations previously defined. This indicates that the stations in Group 4, which are predominantly in low job density areas, pose a challenge for the model.

We learn from the analysis results that spatial and temporal patterns can effectively complement each other, contributing to the final prediction result. Without clear delineation, these two features build upon each other and influence the model. However, a preliminary dataset clustering before training, especially for system-level training, could be beneficial to avoid interference from different feature groups.

\section{Conclusion}\label{sec:conc}
In this work, we proposed an innovative Hybrid Deep Learning (DL) model that effectively combines dynamic spatial and temporal feature extraction into a unified framework, capable of performing network-wide station-level ridership prediction. The model outperformed other Machine Learning (ML) and DL models across three distinct prediction scenarios, each with varying features. Our model demonstrated excellent accuracy and robustness, maintaining relatively steady performance throughout different times of the day and across different prediction scenarios. Additionally, it showed consistent performance in long-term predictions using a simple iterative prediction framework, which is both highly efficient and applicable.

Furthermore, we conducted a detailed analysis of results from both temporal and spatial perspectives to provide a reference for future prediction system design. This comprehensive analysis helps to better understand the strengths and potential improvements of the proposed model.

For future work, we aim to apply the model to solve real-world tasks, such as abnormal situation prediction or system monitoring, leveraging its stable performance in long-term predictions with fine-grained resolution. We also plan to explore better solutions for long-term predictions, minimizing performance deterioration over extended prediction intervals.

% \section*{REFERENCES}

\bibliographystyle{IEEEtran}
\bibliography{references}

\end{document}